\documentclass{article}

\usepackage[main,preprint]{mystyle}
\usepackage[utf8]{inputenc}
\usepackage[T1]{fontenc}
\usepackage{natbib}
\usepackage{microtype}
\usepackage{booktabs}
\usepackage{graphicx}\usepackage{float}
\usepackage{svg}
\usepackage{subcaption}
\usepackage{amsmath,amssymb,amsfonts,mathtools}
\usepackage{multirow}
\usepackage{adjustbox}
\usepackage{enumitem}
\usepackage[table]{xcolor}
\usepackage[most]{tcolorbox}
\usepackage{ragged2e}
\usepackage{array}
\usepackage{hyperref}
\usepackage{url}

\usepackage{booktabs}
\usepackage{adjustbox}

\usepackage{amsmath,amsfonts,bm}

\def\eqref#1{equation~\ref{#1}}

\def\1{\bm{1}}

\DeclareMathAlphabet{\mathsfit}{\encodingdefault}{\sfdefault}{m}{sl}
\SetMathAlphabet{\mathsfit}{bold}{\encodingdefault}{\sfdefault}{bx}{n}

\setlist[itemize]{leftmargin=*}
\setlist[enumerate]{leftmargin=*}

\title{%
  \raisebox{-0.10cm}{\includegraphics[height=0.65cm]{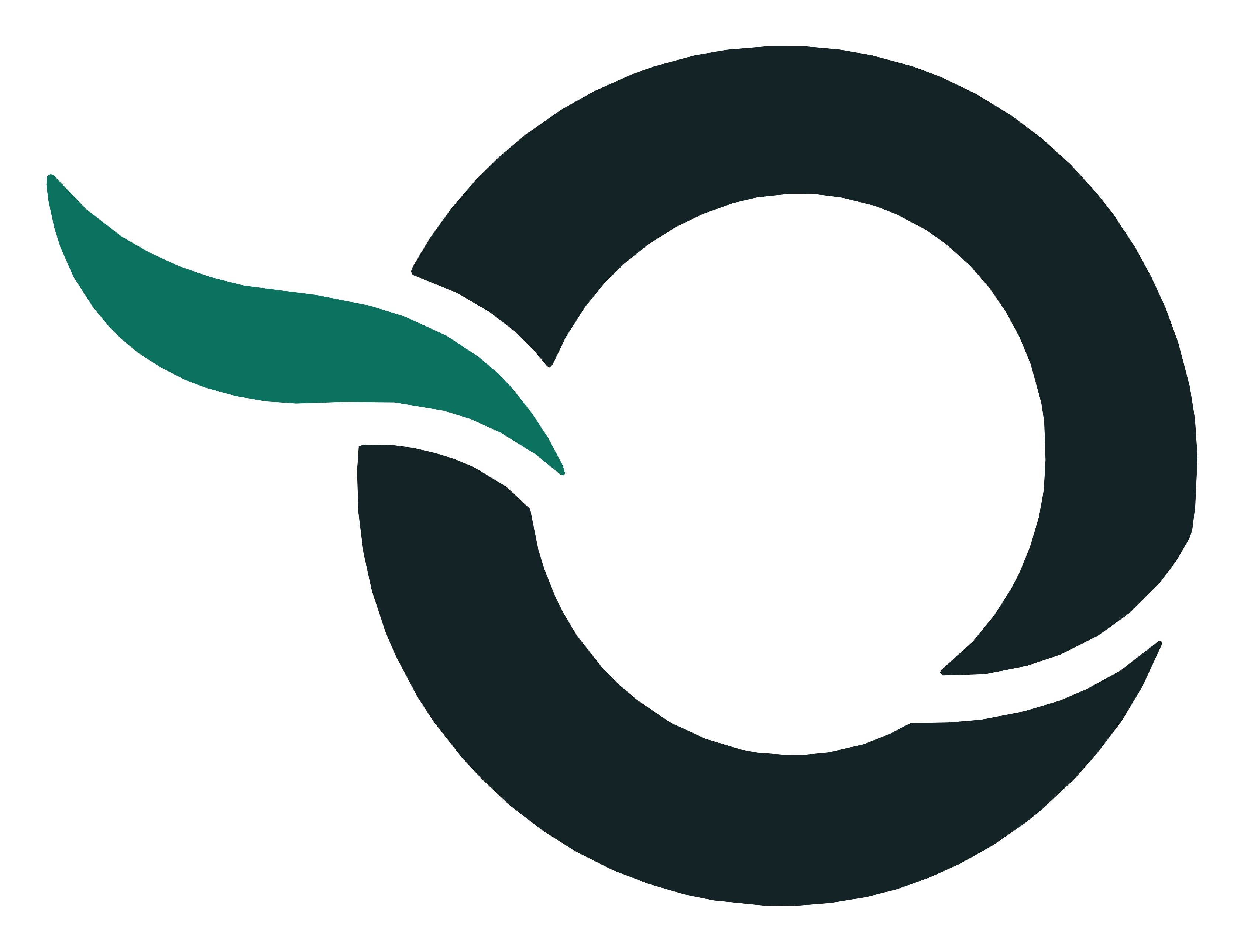}}\hspace{-0.02cm}ccamy-1.0:\\
  Open Pareto-frontier 35B Intelligence for Co-work}

\author{\textbf{Accio Team}}

\headerlogos{%
  \includegraphics[height=0.45cm]{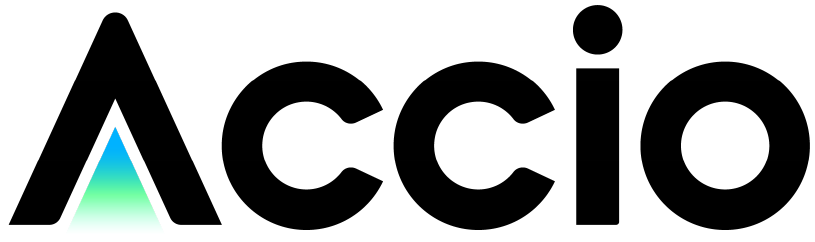}%
}

\begin{document}

\maketitle

\begin{abstract}
Co-work agents execute complex workflows that combine information gathering, tool use, coding, and file manipulation across many model invocations. Because cost and latency accumulate over the full episode, their practical value depends not only on peak capability but also on how efficiently that capability is delivered. Yet many steps in everyday work emphasize state tracking, coordination, recovery, and follow-through rather than frontier-scale reasoning. We present Occamy-1.0, a cost-efficient co-work model obtained by further training the post-trained Qwen3.6-35B-A3B checkpoint. We construct execution-grounded data and environments, capture replayable long-horizon trajectories across multiple harnesses, and use staged post-training to develop and consolidate complementary execution capabilities. Across a broad suite of co-work benchmarks, Occamy-1.0 is consistently among the strongest comparably sized models and remains competitive with substantially larger frontier systems on several tasks. Under our stated evaluation and pricing protocol, its aggregate performance across four representative benchmarks places it at the low-cost knee of the observed cost--performance Pareto frontier. Supporting evaluations in tool calling, coding, and instruction following further show that this specialization preserves broad agentic capability. We release the model weights and a subset of the training data to support research on practical co-work agents and agentic post-training.\footnotemark\end{abstract}\footnotetext{\begin{tabular}[t]{@{}ll@{}}
& \url{https://huggingface.co/Accio-Lab/Occamy-1.0}\\[0.15em]
& \url{https://github.com/Accio-Lab/occamy}
\end{tabular}}
\vfill\begin{center}  \includegraphics[width=\linewidth]{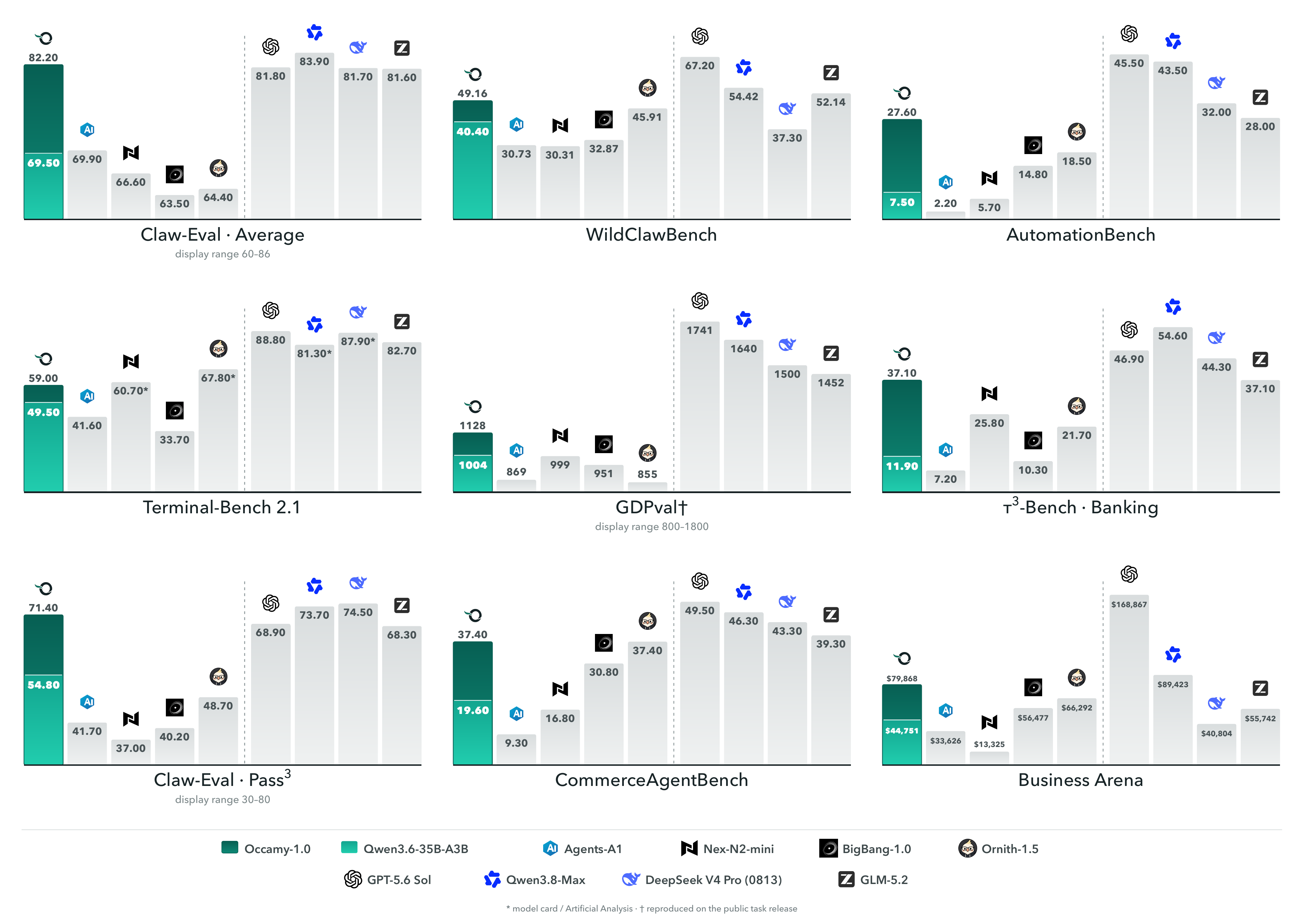}\end{center}\vspace{-0.8em}\vfill
\clearpage
\tableofcontents
\clearpage

\section{Introduction}
\label{sec:introduction}

\begin{figure}[H]
\centering
\includegraphics[width=\linewidth]{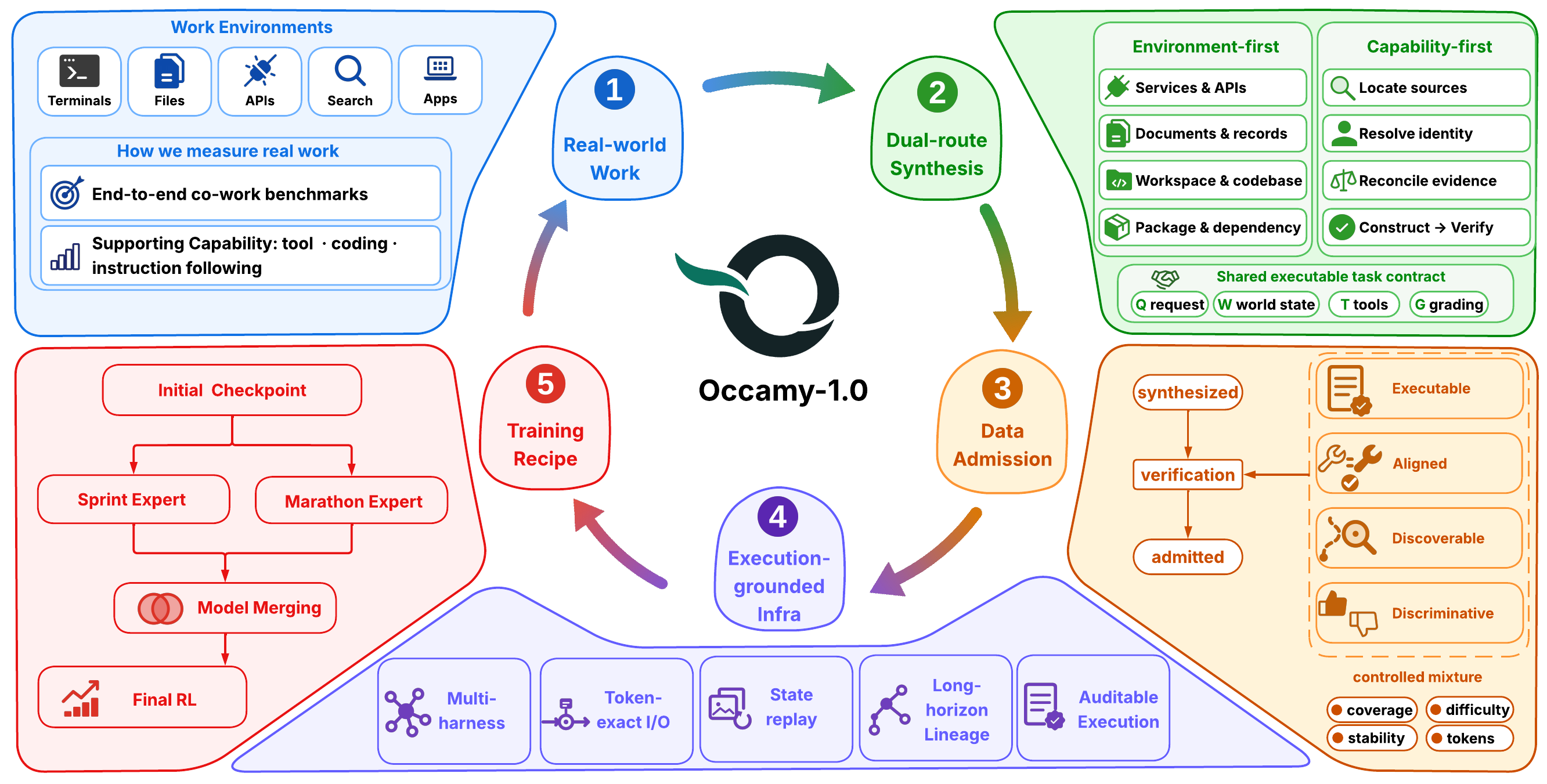}
\caption{Overview of Occamy-1.0. The system connects real-work task construction, verified data admission, replayable multi-harness execution, and staged post-training.}
\label{fig:overview}
\end{figure}

\paragraph{Co-work is an end-to-end workload.}
Digital agents are increasingly asked to carry out real work rather than answer isolated questions: update CRM records, complete finance workflows, operate an e-commerce business, or handle the everyday tasks involved in running a company. These settings combine dense context---customer histories, documents, policies, transactions, and prior decisions---with specialized tools and evolving external state. An agent may need to gather information, write and run code, edit files, invoke structured tools, inspect intermediate results, and recover from failed actions. We use \emph{co-work} to describe this user-directed, multi-step work in a persistent digital environment. It may draw on coding, information gathering, and tool use, but is defined by sustained coordination across the complete task rather than by any fixed collection of skills.

\paragraph{Why efficiency matters.}
Co-work changes the economics of model inference. A long task can invoke the model dozens or hundreds of times, so small differences in per-call cost and latency accumulate across the episode. At the same time, capability demand is uneven. Some steps require difficult reasoning, while many others depend on accurate state tracking, disciplined tool use, recovery, and reliable follow-through. Using a frontier-scale model for every step can therefore be unnecessarily expensive. The practical objective is not simply the highest standalone benchmark score, but strong end-to-end execution under realistic cost, latency, and deployment constraints. This motivates our central question: can a compact model move the capability--cost Pareto frontier for co-work?

\paragraph{Occamy-1.0.}
We develop Occamy-1.0 by further training the post-trained Qwen3.6-35B-A3B checkpoint~\citep{qwen2026qwen36_35b_a3b}. Because the starting checkpoint already provides strong language, reasoning, and coding capabilities, we focus additional training on turning those capabilities into reliable execution: following tool contracts, acting on observed environment state, recovering from failure, maintaining continuity across history rewrites, and optimizing complete task outcomes. Our goal is not to replace frontier systems on every possible task, but to provide a practical co-work model that delivers a favorable level of capability at substantially lower inference cost and remains feasible to self-host.

\paragraph{Continuity over long horizons.}
The core technical challenge is maintaining continuity throughout a long task. Task state persists across many model invocations and may outgrow a single context window. The execution harness may therefore compact, prune, or rewrite the history visible to the model while the environment continues to evolve. These changes alter what the model sees but do not reset what has already happened: files remain written, tool effects persist, and earlier decisions continue to constrain later actions. A single episode can consequently span multiple model-visible histories while remaining one continuous task. The infrastructure must capture exactly what the policy observed and produced, preserve lineage across history rewrites and delegated runs, and replay environment state faithfully. The learner must then assign task-level outcomes to the correct policy tokens across all invocations and segments.

\paragraph{The Occamy recipe.}
Figure~\ref{fig:overview} summarizes three connected components. First, execution-grounded data and environments connect task construction to runnable tools, observable state transitions, realized trajectories, and task-level grading. Candidate task--environment packages pass through a shared verification and admission process before entering the training mixture. Second, harness-aware infrastructure preserves harness-specific token, tool, rewrite, and state semantics while exposing a common trajectory and replay contract to the learner. Third, staged post-training develops and consolidates complementary execution capabilities. The \emph{Marathon Expert} targets sustained execution through SFT followed by Hierarchical Decoupled Policy Optimization (HDPO)~\citep{yan2026act}. The \emph{Sprint Expert} uses SFT over a broader distribution of shorter-horizon agentic tasks, including coding, information gathering, and structured tool use. We combine the experts through model merging~\citep{wortsman2022modelsoups,yadav2023tiesmerging}, then apply Single-Rollout Asynchronous Optimization (SAO)~\citep{hou2026singlerolloutasynchronousoptimizationagentic} to the merged checkpoint on a broad co-work mixture.

\begin{figure}[t]
\centering
\includegraphics[width=\linewidth]{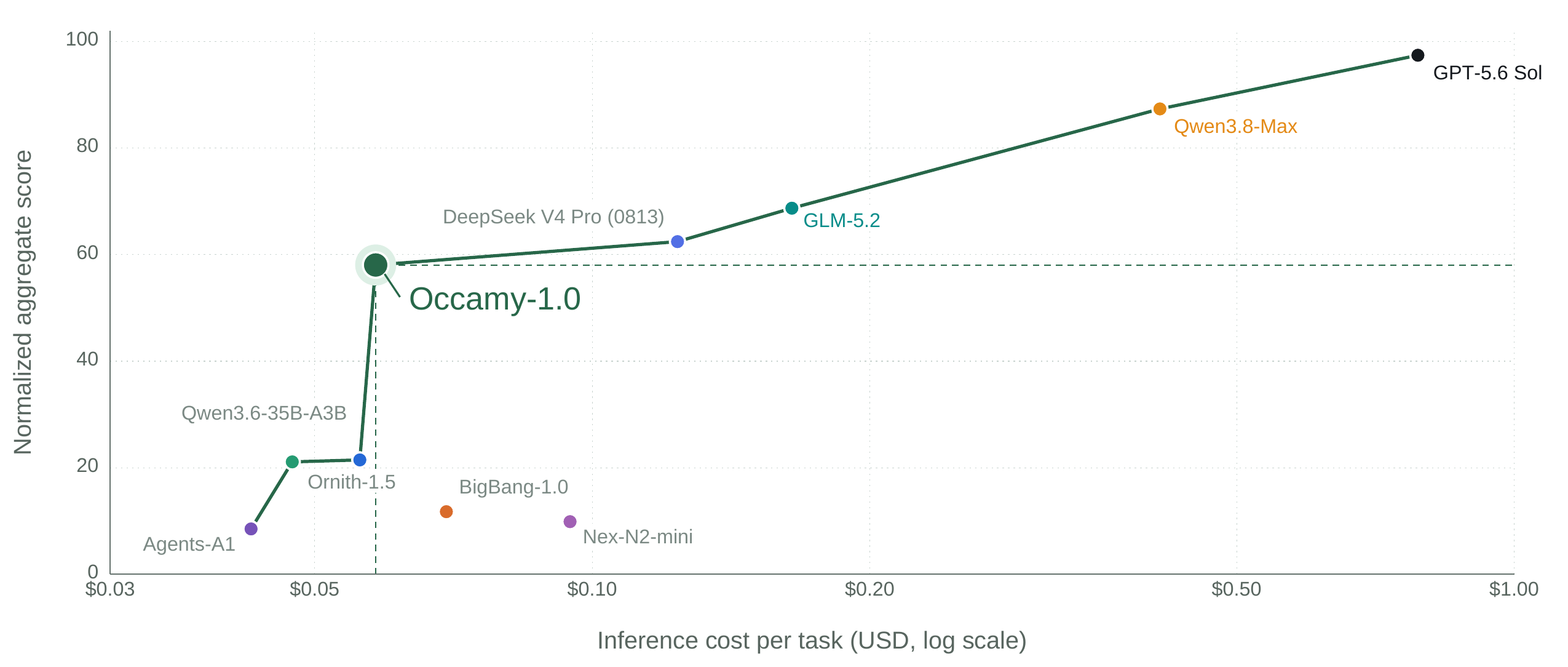}
\caption{Aggregate cost--performance across Claw-Eval, WildClawBench, AutomationBench, and GDPval. The green line traces the empirical Pareto frontier; Occamy-1.0 lies near its low-cost knee. Section~\ref{sec:cost-efficiency} gives the full evaluation and pricing protocol.}
\label{fig:aggregate-cost-performance}
\end{figure}

\paragraph{Results.}
Occamy-1.0 performs consistently near the top of the comparably sized group across the co-work suite, including Claw-Eval~\citep{claweval2026}, WildClawBench~\citep{wildclawbench2026}, CommerceAgentBench~\citep{realreplica2026}, and Business Arena~\citep{businessarena2026}, while remaining competitive with substantially larger systems on several evaluated tasks. Supporting evaluations in tool calling, coding, and instruction following indicate that this specialization does not collapse broader agentic capabilities. Cost efficiency is part of the same result: Figure~\ref{fig:aggregate-cost-performance} shows that aggregate performance across four representative benchmarks places Occamy near the low-cost knee of the observed Pareto frontier under the common protocol detailed in Section~\ref{sec:cost-efficiency}. Relative to its Qwen3.6-35B-A3B starting checkpoint, Occamy delivers a large capability gain with only a modest change in per-task inference cost.

\paragraph{Contributions.}
This report documents the system choices behind this capability--cost operating point:
\begin{itemize}
    \item \textbf{Execution-grounded data and environments.}
    We connect task construction, runnable environments, realized tool behavior, trajectory collection, and task-level grading in one traceable pipeline. Environment-first and capability-first synthesis routes share the same verification and admission process.
    \item \textbf{Harness-aware infrastructure for long episodes.}
    Our adapters preserve harness-specific token, tool, rewrite, and state semantics, while the training layer uses a shared trajectory schema and replay contract. Token-exact capture and environment-state replay preserve trajectory fidelity across history rewrites and heterogeneous harnesses.
    \item \textbf{A specialization-and-consolidation training recipe.}
    We train a \emph{Marathon Expert} for sustained execution using SFT followed by HDPO and a \emph{Sprint Expert} on a broader distribution of shorter-horizon agentic workloads. We combine them through model merging and apply SAO to the merged checkpoint on a broad co-work task mixture.
\end{itemize}

\paragraph{Report roadmap.}
Section~\ref{sec:cowork} defines the co-work setting and the turn-, segment-, and episode-level abstractions used throughout the report. Sections~\ref{sec:data_env}--\ref{sec:posttraining} describe the data and environments, training infrastructure, and post-training recipe. Section~\ref{sec:evaluation} presents the co-work results together with supporting capabilities and efficiency measurements, and Section~\ref{sec:analysis} summarizes practical lessons and remaining limitations.

\section{Co-work Setting and Terminology}
\label{sec:cowork}

Co-work execution is organized around task-scoped episodes. A \textbf{task} fixes the objective, initial environment, available tools, and grading contract; an \textbf{episode} is one attempt under a model and harness and receives one task-level outcome. The episode begins with a root run and may branch into subagent runs, each recorded as its own trajectory. Within a run, turns remain in one segment while the model-visible history grows append-only; a harness-declared summary or pruning operation is a \textbf{history rewrite} that closes the current segment and starts another without changing the run, episode, or shared environment. Policy outputs are trained only where they were sampled; inherited or rewritten context, tool observations, and returned subagent results provide conditioning rather than additional training targets. Table~\ref{tab:cowork-terminology} summarizes these boundaries, while Appendix~\ref{app:episode-anatomy} illustrates the complete structure.

\begin{center}\begin{minipage}{\linewidth}
\centering
\small
\setlength{\tabcolsep}{4pt}
\renewcommand{\arraystretch}{1.08}
\captionsetup{hypcap=false}\captionof{table}{Terminology used throughout the report.}
\label{tab:cowork-terminology}
\begin{tabular}{@{}p{0.14\linewidth}p{0.48\linewidth}p{0.30\linewidth}@{}}
\toprule
Term & Operational definition & Boundary rule \\
\midrule
Conversation &
The user-facing interaction container. &
May contain multiple tasks and task-scoped episodes. \\
Task &
A fixed objective, initial environment, tool set, and grading contract. &
May be attempted by multiple episodes. \\
Episode &
One attempt to execute one task under a particular model and harness. &
Receives one task-level outcome. \\
Agent run &
One independently stepped policy loop with its own model-visible history. &
An episode has one root run and may contain child runs. \\
Subagent run &
A child run spawned by a parent run to handle a delegated subgoal. &
Creates a new trajectory, but remains in the same episode; its result returns as a parent observation. \\
Trajectory &
The recorded policy inputs, outputs, and execution events of one agent run. &
Maps to one run and may contain multiple segments. \\
Segment &
An append-only span of model-visible history within a trajectory. &
A declared history rewrite closes it and starts the next segment. \\
Turn &
One model-visible input, its sampled policy output, and the resulting tool events. &
The atomic recorded interaction within a segment. \\
History rewrite &
A harness-declared transformation of an existing model-visible prefix. &
Starts a new segment while the run, environment, and episode continue. \\
\bottomrule
\end{tabular}
\end{minipage}\end{center}

\section{Data and Environments}
\label{sec:data_env}
In this section, we introduce the construction of tasks and environments, a
major challenge in building co-work agent data. In particular, realistic and
reproducible environments are essential for both SFT trajectory collection and
RL policy rollouts. To address this challenge, we construct co-work data under
a common set of task-design principles. These
principles govern how the request, world state, tools and services, and
grading contract are specified and connected, so that each generated task is
grounded, feasible, and verifiable under controlled information boundaries.
Guided by these principles, we develop two complementary task-synthesis
routes: \textbf{environment-first synthesis} and \textbf{capability-first synthesis}. Each
pipeline applies construction-time checks suited to its generation process,
after which all tasks pass through shared canonicalization, verification, and
training-admission gates.

\subsection{Executable Task Contracts}
\label{subsec:task_environment_spec}

We represent each reusable task as a versioned executable contract consisting
of two components: a \emph{task specification} and \emph{orchestration
metadata}. The task specification comprises four elements: the \emph{public
request}, the \emph{initial world state}, the \emph{tool specification}, which
defines the permitted tools and their state transitions, and the \emph{private
completion and grading contract}. The orchestration metadata records stable
identity, schema version, required capabilities, deterministic time settings,
and execution budgets.

These components must be jointly consistent: the evidence required by the
public request must be present in the initial world state and reachable through
the permitted tools; every grading condition must be justified by the request
and observable from execution evidence; and the specified tool contracts must
faithfully match their backends. Because these relationships concern task
semantics and executability, static schema validation alone is insufficient.
An infeasible task can appear indistinguishable from a difficult one when both
produce a low score. Our synthesis and admission pipeline therefore requires
execution-level evidence both that a valid completion path exists and that the
grader distinguishes correct work from no-ops and other degenerate strategies.

\subsection{Task Design Principles}
\label{subsec:task_validity}

Building on the above contract-level consistency requirements, both synthesis routes follow a common set of task-design principles, stated in
full in Appendix~\ref{app:task_design_principles}. Together, these principles require
each task to be framed as a realistic professional commission whose difficulty
comes from the work itself rather than from ambiguity, and to be solvable with
exactly the information and tool access available to the evaluated policy. Tool contracts must faithfully describe
their backends, and grading is applied to sealed execution evidence under a
contract that is frozen before any evaluated rollout and hidden from the policy. These shared principles provide the foundation for the task-synthesis
procedures described below.

\subsection{Co-work Capability Space and Task Variation}
\label{subsec:capability_space}

\begin{figure}[t]
\centering
\includegraphics[width=\linewidth]{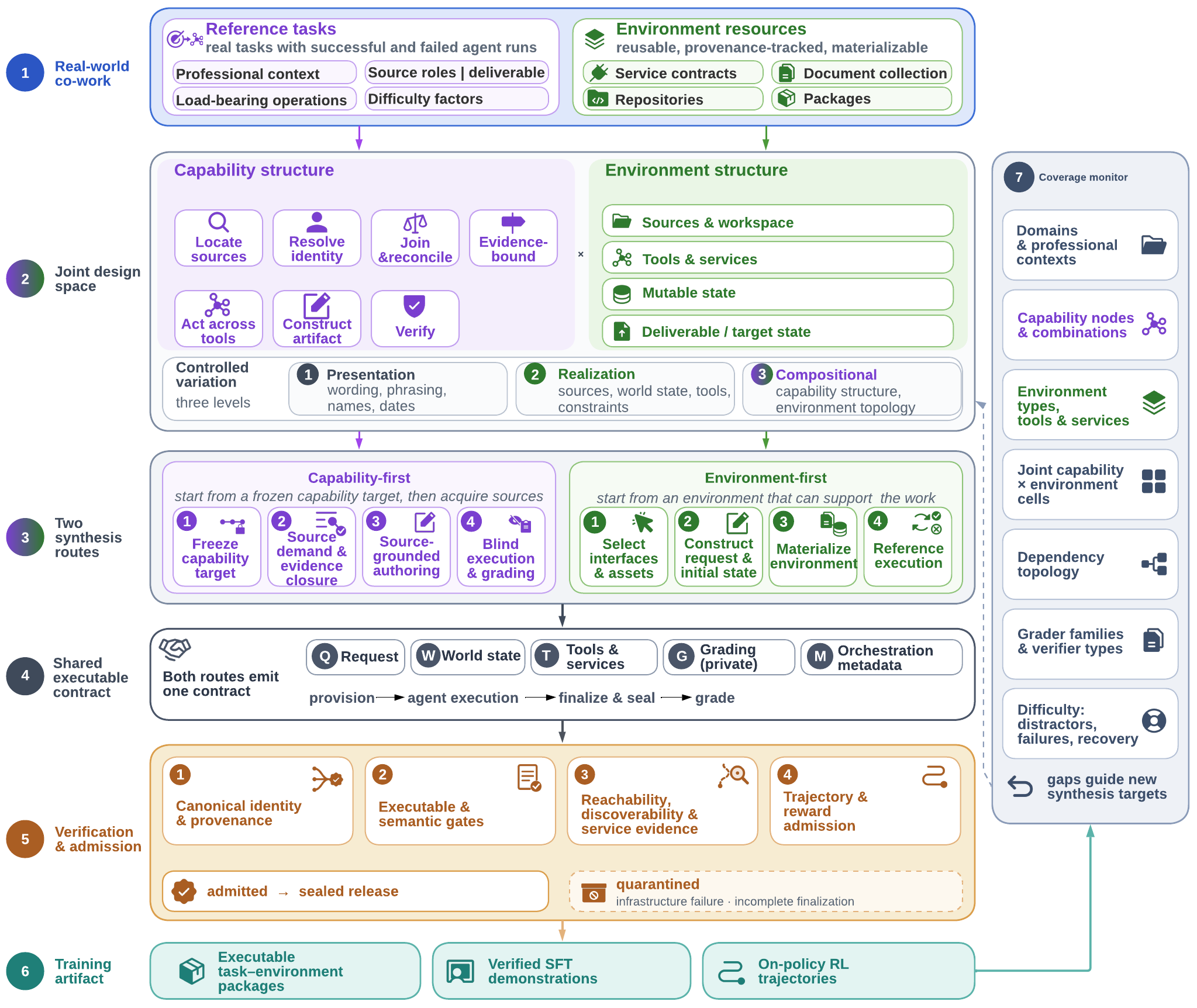}
\caption{From real-world work to execution-grounded training data. Real co-work
commissions and reusable environment resources~(1) are abstracted into the joint
capability--environment design space and its three levels of controlled
variation~(2). Two complementary synthesis routes~(3) emit the same versioned
executable contract $\mathcal{C}=(\mathcal{S},\mathcal{M})$ and its execution
lifecycle~(4), which is admitted through one shared canonicalization,
verification, and admission pipeline~(5) into three separately counted training
artifacts~(6). Collection-level coverage statistics~(7) feed back into the
synthesis targets.}
\label{fig:data-overview}
\end{figure}

Our real-world co-work task collection and environment resources provide
authentic grounding but yield too few complete executable contracts for
training at scale. To organize scalable task construction, we situate each task in a joint capability--environment space, as shown in block~(2) of Figure~\ref{fig:data-overview}.
A capability structure records the transferable operations required by the task
and the load-bearing dependencies among them. An environment structure records
the evidence carriers, mutable state, action interfaces, and verifiable outcomes
through which those operations are instantiated. Detailed definitions are provided in
Appendix~\ref{app:capability_environment_structures}.

Within this shared space, we construct additional tasks from two complementary
starting points: environment-first synthesis starts from real
working environments, including documents, repositories, software packages,
and service affordances, and constructs tasks that those environments can
support and verify. Capability-first
synthesis begins with capability dependencies abstracted from real work, and then acquires independent sources
and environments in which the target capabilities can be instantiated in new
tasks. Figure~\ref{fig:data-overview} summarizes both routes, the shared executable contracts they produce, the corresponding execution instances, and
the common admission pipeline. Detailed construction procedures and
route-specific gates are provided in
Appendix~\ref{subsec:environment_first_synthesis} and
Appendix~\ref{subsec:capability_first_synthesis}.

Beyond constructing additional tasks, scaling the collection requires
controlling how newly constructed tasks vary across the joint
capability--environment space defined above. We organize task variation into
three levels of increasing structural change.
\emph{Presentation variation}
changes surface attributes such as language, persona, names, or dates while
leaving the underlying capability demands and environment topology unchanged.
\emph{Realization variation} preserves the target capabilities and their broad
dependency structure while changing the environment realization, including its
sources, initial state, tools or services, constraints, and deliverables.
\emph{Compositional variation} changes the capability structure, the
environment topology, or the coupling between them, creating new dependencies
among evidence, decisions, actions, and verification. We control
collection-level coverage over both structures and their combinations.
Figure~\ref{fig:data-overview} places the three levels within that joint
space; detailed coverage control requirements are provided
in Appendix~\ref{app:task_variation_coverage}.

\subsection{Verification, Admission, and Mixture Control}
\label{subsec:data_admission_mixture}

The two synthesis routes converge on a shared canonicalization and admission
procedure, shown in block~(5) of Figure~\ref{fig:data-overview}. Each
task--environment package is assigned immutable provenance and
runtime identity, and is then checked for executable feasibility, semantic alignment,
discoverability, and grader discrimination. Public-interface reference
executions establish a feasible completion path, while negative executions test
that invalid outcomes receive lower grades. An episode becomes training-eligible
only when its task release, recorded environment effects, and frozen grade can
be joined under a single lineage; incomplete evidence, ambiguous outcome--grade relationships, and
infrastructure failures are quarantined. The complete verification and
admission procedure is provided in
Appendix~\ref{app:canonicalization_admission}.

Admission produces several distinct data artifacts. We account separately for
executable task--environment packages, verified SFT demonstrations, and
on-policy RL trajectories (block~(6) of Figure~\ref{fig:data-overview}), and
assemble the training mixture under joint capability--environment coverage,
difficulty, stability, and trainable-token constraints; block~(7) lists the
coverage axes we monitor at the collection level. Sealed internal release manifests retain
raw, verified, admitted, and quarantined counts for every source. Appendix~\ref{app:data_mixture_coverage} describes the
mixture units, accounting rules, and coverage controls.

\section{Agent Execution Infrastructure}
\label{sec:infra}

\subsection{System Overview}

\paragraph{From live work to training records.}
A co-work episode touches three systems. The harness decides what the model sees and how tools are called. The environment carries files, service state, and the effects of earlier actions. The learning backend needs the exact tokens that the policy sampled. A chat transcript captures only part of this process: one episode may contain several agent runs, each run may cross history rewrites, and the final outcome depends on state that never appears fully in text.

\begin{figure}[t]
\centering
\includegraphics[width=\linewidth]{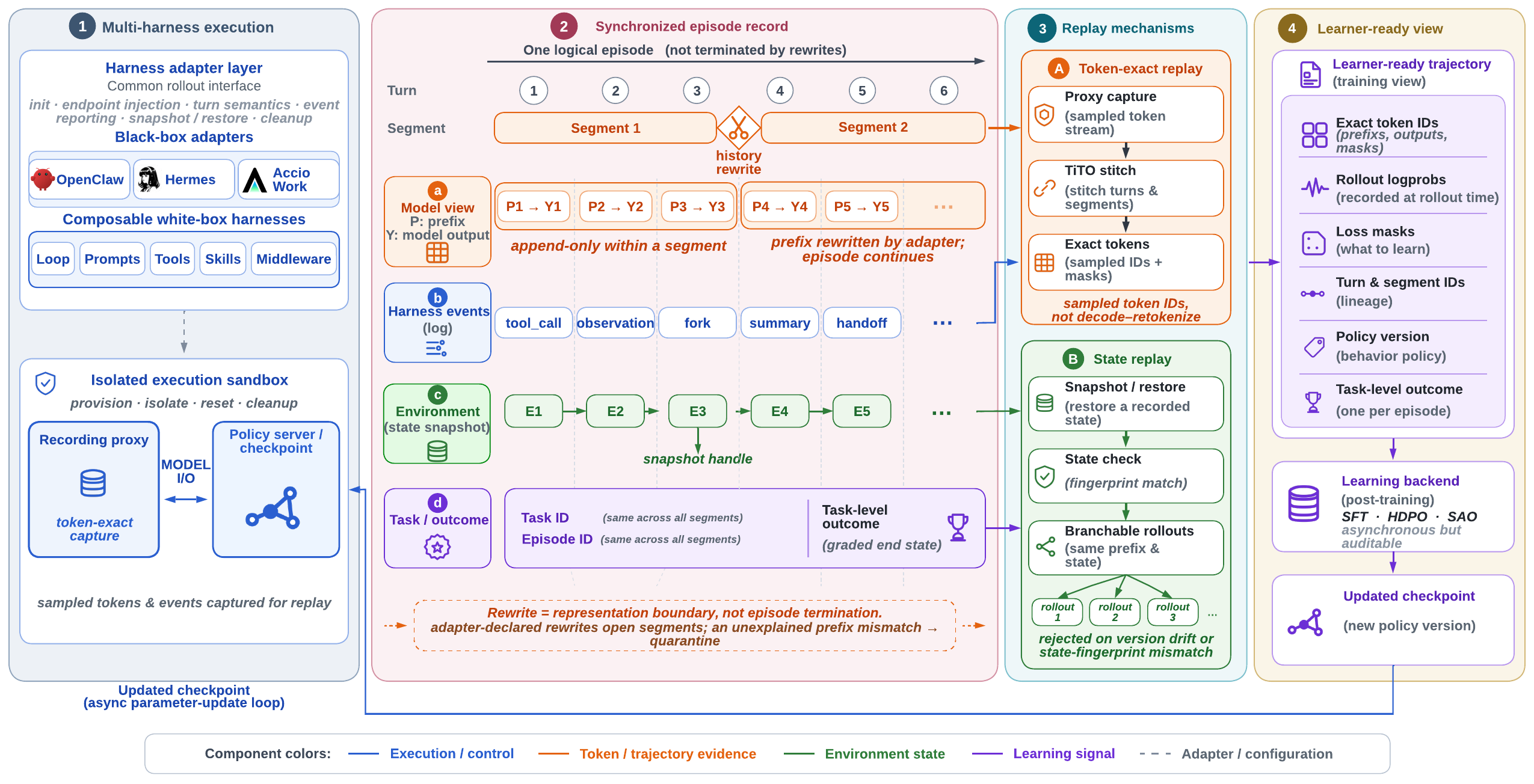}
\caption{Training infrastructure for long-horizon co-work. The four blocks connect multi-harness execution, a synchronized episode record, token- and state-level replay, and the learner-ready training view under one lineage.}
\label{fig:infra-overview}
\end{figure}

\paragraph{One lineage from task to outcome.}
Our infrastructure starts with a frozen task and ends with a sealed episode record. It is harness-aware at the adapter layer and harness-agnostic at the training layer. Adapters preserve harness-specific prompt, tool, rewrite, and session semantics; the learning backend consumes one shared trajectory schema and replay contract. Each episode is bound to its harness, environment version, model checkpoint, grader, and execution budget. It enters training only when the policy view, environment effects, and grader outcome can be joined under the same episode lineage. Figure~\ref{fig:infra-overview} summarizes this path in four blocks: multi-harness execution, a synchronized episode record, token and state replay, and the learner-ready training view.

The rest of this section follows the same order. An episode moves through provisioning, agent execution, finalization and sealing, and grading. These stages may run asynchronously, but their identifiers and evidence remain auditable end to end.

\subsection{Multi-Harness Execution}

\paragraph{Different harnesses, one training contract.}
Harnesses disagree on more than formatting. They use different prompts, tool schemas, turn boundaries, compaction rules, subagent handoffs, timeout behavior, and session lifecycles. We keep these choices inside adapters instead of flattening them into text. The adapter presents a common episode lifecycle while preserving the semantics that produced each turn, action, and rewrite. In this work, we run Occamy under three harnesses through this adapter layer:
OpenClaw~\citep{openclaw}, Hermes~\citep{hermes}, and Accio Work.

\paragraph{Composable white-box harnesses.}
A white-box harness is assembled from reusable parts rather than implemented as a monolithic runtime:
\[
h=\operatorname{Compose}\!\left(\ell,p,\mathcal{U},\mathcal{K},m_{1:k}\right),
\]
where $\ell$ is the agent loop, $p$ its prompt configuration, $\mathcal{U}$ the exposed tool set, $\mathcal{K}$ a bundle of reusable skills, and $m_{1:k}$ an ordered middleware chain. Middleware handles context compaction, retry and recovery, budget enforcement, failure normalization, and event tracing. All components share the same turn, tool-event, history-rewrite, and termination contracts, so one part can change without changing the task specification or downstream trajectory format.

\paragraph{Scaling through structured variation.}
This factorization is our primary mechanism for white-box harness scaling. Before a rollout, the controller resolves the composition, checks tool requirements, rewrite support, and state ownership, and stores an immutable composition manifest in the episode lineage. The sampler can then vary individual components or complete compositions while holding the task and grading specification fixed. Training across this structured variation reduces dependence on any single harness realization and improves transfer to compositionally related harnesses.

These two sources of variation serve different purposes. Multi-harness training asks the policy to solve a task across prompt layouts, tool schemas, rewrite rules, and recovery conventions rather than fit one runtime. White-box composition adds control and observability: we can vary one component at a time, replay the same work state, and attribute failures to the loop, tool, skill, or middleware choice. Together, they turn harness diversity into a controlled training variable that supports both transfer and targeted data generation.

\paragraph{Black-box, white-box, and delegated runs.}
A black-box adapter runs an external agent process behind a service boundary, while a resolved white-box composition controls the tool loop directly. Both expose the same initialization, execution, event, replay, and finalization contract. The same frozen task can therefore be attempted under different harnesses without changing its grading contract, giving the policy experience with different execution and recovery conventions.

Delegation follows the same rule. A child run keeps the parent episode, fork point, and handoff event, but owns separate session and trajectory identifiers. Its result returns to the parent as an observation rather than being concatenated into the parent trajectory. This preserves the trajectory and loss-accounting boundaries defined in Section~\ref{sec:cowork}.

\subsection{Token-Exact Trajectory Capture and Replay}

\paragraph{Capture at the model boundary.}
A chat log is not an exact policy record. Decoding and re-encoding can change token IDs, a chat template can re-render earlier messages, and a harness can omit internal content. We therefore route every policy call through an OpenAI-compatible proxy. The proxy records input and sampled token IDs, rollout log-probabilities when required, loss masks, model version, and the owning episode, run, turn, and segment.

\paragraph{Replay without retokenization.}
Within an append-only segment, Token-In-Token-Out (TiTO) preserves sampled IDs
and tokenizes only newly introduced non-policy content. We define turn
$u_t=(P_t,Y_t,D_t)$, where $P_t$ is the exact token-ID prefix supplied to the
model, $Y_t$ the policy-output tokens sampled from it, and $D_t$ the
model-visible non-policy content introduced before the next policy call (tool
results, observations, and harness-inserted messages). The invariant is
\[
P_{t+1}=\operatorname{Append}_h(P_t,Y_t,D_t),
\]
at the token level, not merely after decoding, where $\operatorname{Append}_h$
is the deterministic rendering of harness $h$ (chat template, tool
serialization, and role formatting). The stored fragments are stitched into the exact sequence, log-probabilities, policy versions, and loss mask consumed by training. Stable episode, run, and turn identifiers also make network retries idempotent rather than duplicate actions.

\paragraph{Make rewrites explicit.}
We admit two adapter-declared prefix transformations: summary replacement and history pruning. Either transformation closes the current segment and restarts TiTO from the new model-visible prefix; append-only history does not. The record preserves the rewrite type, the prefixes before and after it, and the provenance of any inserted summary. An unexplained prefix change, or a failure in tokenization, rendering, or token capture, keeps the episode available for audit but excludes it from training.

\subsection{State Replay and Episode Finalization}

\paragraph{Replay the work state.}
Tokens tell only half of the story. State replay reconstructs the files, services, tools, and session state on which an action operated. An adapter may restore a snapshot, rebuild a deterministic fixture and replay the action prefix, or resume through a stable session handle. A replay is admitted only when the relevant fingerprints and grader preconditions match. Hidden grader state, restore handles, and future information remain outside the model-visible prefix.

This makes it possible to branch several policy samples from the same prefix and work state, train recovery from difficult intermediate states, and compare checkpoints without rerunning an entire episode. Environment drift, missing dependencies, non-idempotent reconstruction, or a fingerprint mismatch causes rejection. Table~\ref{tab:replay-contracts} summarizes the replay mechanism and validation check at each layer.

\begin{table}[t]
\centering
\small
\caption{Replay and reconstruction mechanisms across a co-work episode.}
\label{tab:replay-contracts}
\begin{tabular}{p{0.14\linewidth}p{0.27\linewidth}p{0.27\linewidth}p{0.21\linewidth}}
\toprule
Layer & Replayed object & Mechanism & Validation \\
\midrule
Policy & Prefix and response tokens, log-probabilities, masks & Proxy capture and TiTO stitching & Token-prefix and length equality \\
Run lineage & Turns, segments, rewrites, forks, handoffs & Harness events and lineage records & Append-only audit and parent/fork metadata \\
Environment & Workspace, service, tool, and session state & Snapshot/restore or action replay & State fingerprints and grader preconditions \\
Episode & Artifacts, outcome evidence, completion status & Versioned grader replay & Outcome and artifact consistency \\
\bottomrule
\end{tabular}
\end{table}

\paragraph{Seal once, grade once.}
Execution state is isolated by episode and remains independent of harness choice. After the root run terminates and all child runs are finalized, the system joins their token records with harness events, artifacts, and final environment evidence. The record is sealed before grading, and the grader attaches one episode-level outcome. Each run retains a distinct episode identifier, while all segments and evidence remain linked to the same episode and preserve the parent--child relationships among runs. 

Token mismatch, state drift, incomplete lineage, and infrastructure failure quarantine an episode rather than produce a policy outcome. Reliability reporting separates model failure from harness, proxy, sandbox, environment, and grader failure.

\subsection{Open-Source Release}
Dressage~\citep{dressage_github} is the open-source release of the internal agentic RL infrastructure used to train Occamy. It makes the core training path described in this section public, including multi-harness execution, proxy-mediated trajectory capture, sandbox integration, and multi-segment conversion for RL. The source code and training recipes are available at \url{https://github.com/Accio-Lab/Dressage}. Deployment-specific services and configurations used in this training run are not necessarily included in the public release.

\section{Training Recipe of Occamy}
\label{sec:posttraining}

Occamy follows a staged, multi-branch post-training recipe rather than a single SFT-to-RL pipeline. Starting from the same post-trained checkpoint, we develop two specialized checkpoints: a \emph{Marathon Expert}, trained on long-horizon tasks using SFT followed by HDPO~\citep{yan2026act}, and a \emph{Sprint Expert}, trained with SFT on a broader distribution of shorter-horizon agentic tasks. Here, \emph{expert} refers to a specialized checkpoint rather than an individual routed expert in the underlying MoE architecture. We combine their complementary capabilities through model merging, then refine the merged policy with SAO~\citep{hou2026singlerolloutasynchronousoptimizationagentic} on a broad mixture of co-work tasks. The following subsections describe each technical component, with additional details on training schedules and checkpoint configurations provided in Appendix~\ref{app:training_hyperparameters}.

\subsection{SFT: Behavioral Initialization}

\paragraph{Data Composition.}
Table~\ref{tab:sft_data} reports the deduplicated union of the SFT data used across the Marathon and Sprint experts.  The figures are totals across both experts, not counts for either expert alone.  The combined corpus spans
four domains:

\begin{itemize}
\item \textbf{General agentic:} Multi-turn assistant and computer-use tasks
      over file, web, and media tools, in English and Chinese, together with
      occupational tasks that require producing real professional
      deliverables.  Trajectories cover planning, delegation, evidence
      gathering, and long-form artifact writing.
\item \textbf{Long-horizon interactive agents:} Stateful multi-session
      episodes in simulated environments, requiring sustained planning and
      decision-making across many turns. These are the longest trajectories
      in the corpus designed for model training.
\item \textbf{Terminal and software engineering:} File and command-line work
      in a sandboxed shell, with dense reasoning between actions.
\item \textbf{Tool-call grounding:} Shorter workflow-automation tasks,
      memory-augmented tool selection, and multi-turn service dialogues.  These
      anchor call formatting, argument binding, and policy adherence rather
      than long-horizon planning.
\end{itemize}

In total the corpus contains approximately \textbf{15.0K trajectories} and \textbf{403.3M
tokens}, with an average trajectory length of \textbf{26.9K tokens}.  The
distribution is strongly bimodal: general agentic and long-horizon episodes
supply 72.5\% of the tokens from 42.3\% of the samples, while tool-call
grounding supplies 49.5\% of the samples but only 16.8\% of the tokens.

\begin{table}[ht]
\centering
\caption{Composition of the deduplicated SFT union across both expert training sets.}
\label{tab:sft_data}
\begin{tabular}{lrrrr}
\toprule
\textbf{Data Source} & \textbf{Samples} & \textbf{Share} & \textbf{Avg. Len.} & \textbf{Tokens} \\
\midrule
General agentic                    & 5,418  & 36.1\% & 37.7K & 204.1M \\
Long-horizon interactive agents    & 923    & 6.2\%  & 95.8K & 88.4M  \\
Terminal and software engineering  & 1,228  & 8.2\%  & 35.1K & 43.1M  \\
Tool-call grounding                & 7,429  & 49.5\% & 9.1K  & 67.7M  \\
\midrule
\textbf{Overall}                   & \textbf{14,998} & 100\% & \textbf{26.9K} & \textbf{403.3M} \\
\bottomrule
\end{tabular}
\end{table}

\paragraph{Training Configuration.}
We fine-tune all parameters of the language backbone with cross-entropy loss on
response tokens only; the vision encoder and its projector are frozen.
Hyperparameters are listed in Table~\ref{tab:sft_hparams}.  Reasoning content is
preserved in context across turns rather than stripped between rounds.  Many of
our trajectories continue after the harness compacts the context; for these we
apply \textbf{history masking}: the harness-produced summary and the replayed
prefix are kept as conditioning but excluded from the loss, so the model is
supervised only on the actions it must newly produce and never trained to
regenerate machine-written history.  To improve throughput we pack multiple trajectories into
a single sequence up to the context limit using bin-packing, with attention
masks preventing cross-contamination between packed samples; sequences that
exceed the context limit are dropped rather than truncated.

\begin{table}[ht]
\centering
\caption{Hyperparameters for supervised fine-tuning.}
\label{tab:sft_hparams}
\begin{tabular}{ll}
\toprule
\textbf{Hyperparameter} & \textbf{Value} \\
\midrule
Learning rate            & $1\times10^{-5}$ \\
LR schedule              & Cosine, min $2\times10^{-7}$ \\
Warmup ratio             & 0.05 \\
Optimizer                & AdamW ($\beta_1{=}0.9$, $\beta_2{=}0.95$) \\
Weight decay             & 0.1 \\
Global batch size        & 8 \\
Epochs                   & 5 \\
Max sequence length      & 131,072 \\
Parallelism              & CP=8, EP=8 \\
MoE aux-loss coefficient & $1\times10^{-3}$ \\
\bottomrule
\end{tabular}
\end{table}

\subsection{Reinforcement Learning for Co-work}

We apply reinforcement learning at two stages of the training recipe: first to train the Marathon Expert on long-horizon tasks, and later to refine the merged checkpoint on a broader distribution of co-work tasks. 
In both stages, RL training is defined at the episode level. Each episode \(e\) contains a collection of trajectories, each of which may be partitioned into multiple segments, and receives exactly one terminal task outcome with reward \(R(e)\). A rewrite closes the current segment and begins a new segment within the same episode, but does not produce a new outcome. Let \(\tau\) denote the trajectory used for policy optimization, \(S_{\tau,k}\) the set of model turns in its \(k\)-th segment, and \(Y_t\) the token positions sampled by the policy in turn \(t\). We define the sampled token indices in \(\tau\) as
\[
\mathcal{I}_{\tau}(e)
=
\left\{
(t,i)
\,\middle|\,
t\in\bigcup_{k=0}^{K_{\tau}} S_{\tau,k},
\ i\in Y_t
\right\}.
\]
The policy loss for episode \(e\) is
\[
\mathcal{L}_{\pi}(e)
=
-\widehat{A}_{\mathrm{alg}}(e)
\sum_{(t,i)\in\mathcal{I}_{\tau}(e)}
m_{t,i}\,w_{t,i}
\log \pi_{\theta}
\left(
y_{t,i}\mid P_t,y_{t,<i}
\right).
\]
Here, \(m_{t,i}\in\{0,1\}\) determines whether a sampled token is eligible for training, \(w_{t,i}\) is the algorithm-specific policy-ratio weight, and \(\widehat{A}_{\mathrm{alg}}(e)\) is the episode-level advantage derived from \(R(e)\), together with any baseline or group statistics required by the optimizer. Other trajectories in the episode, including those generated by subagents, remain part of the episode lineage but are excluded from \(\mathcal{I}_{\tau}(e)\) and receive no direct policy loss under the current configuration. Terminal success or failure is recorded exactly once as \(R(e)\), regardless of the number of trajectories or segments in the episode. 

\paragraph{Marathon Expert optimization.}
The Marathon Expert targets at solving long-horizon tasks with multiple turns. Therefore, we additionally conduct RL to improve the efficiency. 
For each task instance, we sample a group of long-horizon interaction trajectories and assign each trajectory a terminal accuracy reward based on the task outcome. We additionally construct an accuracy-conditioned efficiency signal from the number of interaction steps. 
Specifically, we partition each episode group by accuracy reward. For episode $i$ with accuracy reward $R_i^{\mathrm{acc}}$ and
step count $S_i$, let
\[
\mathcal{Q}_i=\{j \mid R_j^{\mathrm{acc}}=R_i^{\mathrm{acc}}\},
\qquad
\overline{S}_i=\frac{1}{|\mathcal{Q}_i|}
\sum_{j\in\mathcal{Q}_i} S_j .
\]
We define its efficiency reward as
\[
R_i^{\mathrm{eff}}
=
-\frac{S_i-\overline{S}_i}{\overline{S}_i}
=
\frac{\overline{S}_i-S_i}{\overline{S}_i}.
\]
trajectories that reach the outcome with fewer steps receive a higher efficiency reward, while trajectories requiring more steps receive a lower reward. Conditioning the comparison on equal accuracy prevents short but lower-quality episodes from being favored over more accurate solutions, while providing a zero-centered relative efficiency signal within each accuracy stratum.

\begin{figure}[t]
\centering
\includegraphics[width=0.75\linewidth]{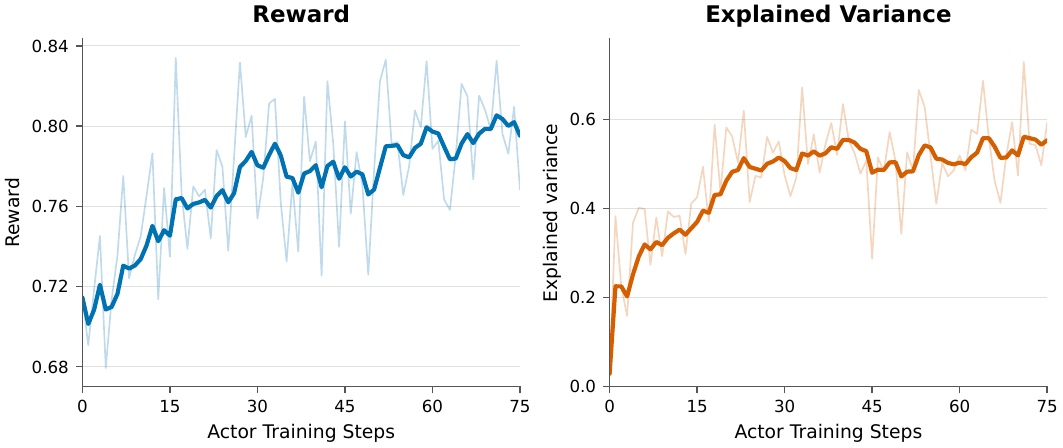}
\caption{Reward and critic explained variance during the final post-merge SAO run.}
\label{fig:reward_explained_variance}
\end{figure}

\paragraph{Final stage co-work RL.}

To further improve the merged checkpoint on co-work tasks, we continue RL training on a broad mixture of co-work tasks.
In this stage, each task instance produces a single interaction trajectory, and completed trajectories are incorporated into training individually. Since the agent trajectories may span multiple context segments, we follow CompactionRL~\citep{li2026compactionrlreinforcementlearningcontext} to continue value estimation and policy optimization across rewrite boundaries. We treat the actual post-rewrite prefix as the next policy state, rather than resetting when the context representation changes. Policy-generated rewrite summaries are trainable under the same objective, whereas harness-produced rewrite content is used only as conditioning (Section~\ref{sec:infra}). Under both optimizers, rollout log-probabilities, masks, and policy-version metadata remain segment-local, while outcomes and lineage remain episode-level. This post-merge stage produces the final Occamy checkpoint. Its training dynamics are shown in Figure~\ref{fig:reward_explained_variance}, with training schedules and hyperparameters provided in Appendix~\ref{app:training_hyperparameters}.

\subsection{Capability Harmonization with Model Merging}

The Marathon and Sprint experts emphasize complementary behavior. We
consolidate them into one deployable checkpoint through parameter-space merging.
Unlike inference-time ensembling, the merged checkpoint has the same parameter count
and serving cost as either parent, and requires no changes to the serving stack.

Both experts share the same architecture, tokenizer, and checkpoint lineage, so
corresponding tensors remain closely aligned and can be averaged directly. We use
a uniform model soup, which requires neither retraining nor inference-time routing.

	Implementation details are summarized here. Equivalent MoE tensors can be serialized under different layouts across the two parent checkpoints, so we normalize them to a common representation before averaging, validate tensor correspondences, accumulate in float32, and independently verify the written shards. Appendix~\ref{app:merge_implementation} provides the full procedure. We also evaluated SLERP~\citep{digitous2023llmslerp}. Under our evaluation setting, it produced similar Claw-Eval performance but a lower BFCL v4 score than the uniform model soup. We use it here only to motivate the merge choice: the uniform model soup improves AutomationBench and Terminal-Bench 2.1 over both parents, keeps WildClawBench between the two experts, and introduces a modest Claw-Eval regression relative to the stronger parent. We therefore treat merging as a capability-balancing step rather than a uniform improvement.

\section{Evaluation}
\label{sec:evaluation}

\begin{table}[H]
\centering
\caption{Full evaluation results. $^{*}$: numbers are taken from official model cards when available, otherwise from Artificial Analysis (AA); all remaining results are reproduced by us under a unified evaluation harness. $^{\dagger}$: our reproduction on the public task release; protocol details in Appendix~\ref{app:eval_registry}. Bold denotes the best score among similarly sized models; underlining denotes the second-best score.}
\label{tab:eval-at-a-glance}

\setlength{\tabcolsep}{2.2pt}
\renewcommand{\arraystretch}{1.12}

\noindent\resizebox{0.96\textwidth}{!}{%
\begin{tabular}{p{3.6cm}*{6}{c}!{\color[HTML]{B8B8B8}\vrule width 0.6pt}*{4}{c}}

\toprule

& \multicolumn{6}{c!{\color[HTML]{B8B8B8}\vrule width 0.6pt}}%
    {\textbf{35B-A3B Models}}
& \multicolumn{4}{c}{\textbf{Large-scale Models}}
\\
\cmidrule(lr){2-7}
\cmidrule(lr){8-11}

\multicolumn{1}{c}{\textbf{Benchmark}}
& \rotatebox[origin=c]{30}{\textbf{Occamy-1.0}}
& \rotatebox[origin=c]{30}{\shortstack{\textbf{Qwen 3.6}\\\textbf{35B-A3B}}}
& \rotatebox[origin=c]{30}{\textbf{Agents-A1}}
& \rotatebox[origin=c]{30}{\textbf{Nex-N2-mini}}
& \rotatebox[origin=c]{30}{\textbf{BigBang-1.0}}
& \rotatebox[origin=c]{30}{\textbf{Ornith-1.5}}
& \rotatebox[origin=c]{30}{\textbf{GPT-5.6 Sol}}
& \rotatebox[origin=c]{30}{\textbf{Qwen3.8-Max}}
& \rotatebox[origin=c]{30}{\shortstack{\textbf{DeepSeek V4}\\\textbf{Pro (0813)}}}
& \rotatebox[origin=c]{30}{\textbf{GLM-5.2}}
\\
\midrule

\rowcolor[HTML]{DCEFEA}
\multicolumn{11}{l}{\textcolor[HTML]{176B5B}{\textbf{Co-work}}}
\\
Claw-Eval (avg.)
& \textbf{82.20}
& 69.50
& \underline{69.90}
& 66.60
& 63.50
& 64.40
& \underline{81.80}
& \textbf{83.90}
& 81.70
& 81.60
\\

Claw-Eval (pass$^3$)
& \textbf{71.40}
& \underline{54.80}
& 41.70
& 37.00
& 40.20
& 48.70
& 68.90
& \underline{73.70}
& \textbf{74.50}
& 68.30
\\

WildClawBench
& \textbf{49.16}
& 40.40
& 30.73
& 30.31
& 32.87
& \underline{45.91}
& \textbf{67.20}
& \underline{54.42}
& 37.30
& 52.14
\\

CommerceAgentBench
& \textbf{37.40}
& 19.60
& 9.30
& 16.80
& \underline{30.80}
& \textbf{37.40}
& \textbf{49.50}
& \underline{46.30}
& 43.30
& 39.30
\\

Business Arena
& \textbf{\$79,868}
& \$44,751
& \$33,626
& \$13,325
& \$56,477
& \underline{\$66,292}
& \textbf{\$168,867}
& \underline{\$89,423}
& \$40,804
& \$55,742
\\

GDPval$^{\dagger}$
& \textbf{1128}
& \underline{1004}
& 869
& 999
& 951
& 855
& \textbf{1741}
& \underline{1640}
& 1500
& 1452
\\

OfficeQA Pro
& \underline{48.10}
& 39.10
& 23.30
& 46.60
& 43.60
& \textbf{59.40}
& \textbf{74.40}
& \underline{69.20}
& 51.20
& 66.20
\\

$\tau^3$-Bench (Banking)
& \textbf{37.10}
& 11.90
& 7.20
& \underline{25.80}
& 10.30
& 21.70
& \underline{46.90}
& \textbf{54.60}
& 44.30
& 37.10
\\
\midrule

\rowcolor[HTML]{FFF0D6}
\multicolumn{11}{l}{\textcolor[HTML]{936400}{\textbf{Tool calling}}}
\\

Automation (Pass$^1$)
& \textbf{27.60}
& 7.50
& 2.20
& 5.70
& 14.80
& \underline{18.50}
& \textbf{45.50}
& \underline{43.50}
& 32.00
& 28.00
\\

Automation (Partial)
& \textbf{69.10}
& 39.40
& 14.70
& 27.90
& 47.40
& \underline{58.00}
& \textbf{81.20}
& \textbf{81.20}
& 59.70
& \underline{70.00}
\\

BFCL v4
& \underline{65.40}
& 63.19
& 57.23
& 62.81
& 57.86
& \textbf{68.51}
& 64.33
& \textbf{73.65}
& 67.10
& \underline{70.33}
\\

VitaBench
& \underline{41.75}
& 34.25
& 37.00
& 26.25
& \textbf{46.00}
& 40.25
& 46.75
& \underline{52.25}
& \textbf{53.50}
& 43.75
\\

\midrule

\rowcolor[HTML]{E2ECF8}
\multicolumn{11}{l}{\textcolor[HTML]{315F91}{\textbf{Coding}}}
\\

Terminal Bench-2.1
& 59.00
& 49.50
& 41.60
& \underline{60.70}$^{*}$
& 33.70
& \textbf{67.80}$^{*}$
& \textbf{88.80}
& 81.30$^{*}$
& \underline{87.90}$^{*}$
& 82.70
\\

\midrule

\rowcolor[HTML]{EEE7F6}
\multicolumn{11}{l}{\textcolor[HTML]{6B528D}{\textbf{Instruction following}}}
\\

IFEval
& \underline{91.53}
& 86.90
& \textbf{91.60}
& \textbf{91.60}
& 90.50
& 81.80
& \underline{95.00}
& \textbf{95.02}
& 93.74
& 93.89
\\

\bottomrule
\end{tabular}%
}
\end{table}

\subsection{Benchmarks and Evaluation Protocol}

We conduct a broad evaluation across 12 benchmarks covering four complementary capability areas. For co-work and office tasks, we evaluate on Claw-Eval~\citep{claweval2026}, WildClawBench~\citep{wildclawbench2026}, CommerceAgentBench (CAB)~\citep{realreplica2026}, GDPval~\citep{gdpval2025}, Business Arena~\citep{businessarena2026}, OfficeQA Pro~\citep{officeqapro2026}, and $\tau^3$-Bench~\citep{tau3bench2026}. Coding-centered capability is assessed using Terminal-Bench 2.1 (TB 2.1)~\citep{terminalbench2026}. We further measure tool use and workflow automation with AutomationBench~\citep{automationbench2026}, BFCL v4~\citep{bfcl2026}, and VitaBench~\citep{vitabench2025}, and evaluate general instruction following with IFEval~\citep{zhou2023instruction}. Whenever possible, we use complete task sets, pinned or official harnesses, isolated execution environments, and benchmark-native graders while maintaining consistent agent scaffolds and serving configurations across compared models. We report each benchmark's native primary metric and explicitly distinguish replicated or modified protocols, such as GDPval and our $\tau^3$-Bench configuration.

\subsection{Compared Baselines}
We compare Occamy-1.0 against two groups of baselines. The first group contains models at the same 35B-A3B scale, including Qwen 3.6 35B-A3B~\citep{qwen2026qwen36_35b_a3b}, Agents-A1~\citep{bai2026agentsa1}, Nex-N2-mini~\citep{nexn22026}, BigBang-1.0~\citep{bigbang2026}, and Ornith-1.5~\citep{ornith2026}, providing a controlled comparison at a fixed model size. The second group consists of substantially larger models, including GPT-5.6 Sol~\citep{openai2026gpt56syscard}, Qwen3.8-Max~\citep{qwen2026qwen38max}, DeepSeek V4 Pro (0813)~\citep{deepseekai2026v4}, and GLM-5.2~\citep{glm2026glm5}, which serve as high-capability reference points.

\subsection{Main Results on Co-work}
\label{subsec:main-cowork}

The overall results are summarized in Table \ref{tab:eval-at-a-glance}, supporting three observations about Occamy-1.0:

\begin{itemize}
    \item \textbf{Strong general co-work performance.} On Claw-Eval, Occamy-1.0 obtains an average score of 82.2, below Qwen3.8-Max (83.92) and above GPT-5.6 Sol (81.8) and DeepSeek V4 Pro (81.7). Its pass$^3$ score is 71.4: above GPT-5.6 Sol (68.9), but below Qwen3.8-Max (73.68) and DeepSeek V4 Pro (74.5). Its average-to-pass$^3$ gap of 10.8 points is the smallest among the comparably sized models, indicating that the gains come from more repeatable execution rather than occasional successful runs.

    \item \textbf{Competitive open-ended and long-horizon execution.} On WildClawBench, Occamy-1.0 scores 49.16, above the comparable-scale models with reported results and below GPT-5.6 Sol (67.20), Qwen3.8-Max (54.42), and GLM-5.2 (52.14). On Business Arena, Occamy-1.0 reaches a final net worth of \$79{,}868, the best result among the comparable-scale models in Table~\ref{tab:eval-at-a-glance} and third overall. The model remains close to the \$80{,}000 initial capital after an extended simulated business workflow, suggesting sustained planning and resource allocation without implying profitable operation.

    \item \textbf{Competitive knowledge-intensive interaction.} Occamy-1.0 obtains 48.1 on OfficeQA Pro, a competitive but not leading result on knowledge-intensive, long-document question answering. On $\tau^3$-Bench (Banking), it scores 37.1, matching GLM-5.2 and exceeding the other comparable-scale models with reported results. A substantial gap remains relative to the strongest larger models on complex document reasoning and simulated-user interaction.
\end{itemize}

Taken together, the results show a broad but uneven profile: Occamy-1.0 is strongest on sustained co-work execution, preserves useful supporting capabilities, and still has clear headroom on knowledge-intensive and simulated-user tasks.






\subsection{Supporting Capabilities: Tool Calling, Coding, and Instruction Following}
\label{subsec:supporting}

Co-work specialization does not turn Occamy-1.0 into a narrow specialist: in addition to its gains on co-work benchmarks, the final model preserves or improves the supporting capabilities required for reliable general-purpose agents, including tool calling, coding, and instruction following.

\paragraph{Tool calling.}
Occamy-1.0 improves over the starting checkpoint on tool-use benchmarks that are not included in its co-work training pool. On AutomationBench, it achieves a strict pass rate of 27.6 and an average partial-credit reward of 69.1, compared with 7.5 and 39.4, respectively, for Qwen3.6-35B-A3B. This corresponds to gains of 20.1 points in strict task completion and 29.7 points in partial-credit reward. The model also reaches 65.40 on BFCL v4, compared with 63.19 for the starting checkpoint, and obtains 41.75 on VitaBench, a 7.50-point increase from 34.25. These results indicate transfer to structured function calling, cross-application automation, and simulated life-service interactions.

\paragraph{Coding.}
On Terminal-Bench 2.1, Occamy-1.0 scores 59.0, improving from 49.5 for the starting checkpoint. It remains slightly below Nex-N2-mini (60.7) and below Ornith-1.5 (67.8), as well as the larger frontier systems in Table~\ref{tab:eval-at-a-glance}, showing a meaningful gain without making coding the model's primary strength.

\paragraph{Instruction following.}
On IFEval, Occamy-1.0 achieves 91.53, improving by 4.63 points over the starting checkpoint's 86.90 and confirming that its stronger agentic behavior is accompanied by, rather than traded against, precise adherence to explicit output constraints.

\paragraph{Generalization beyond the training distribution.}
These supporting results align with the generalization analysis in Section~\ref{sec:analysis}. The final co-work RL stage uses only the Claw task pool and does not include the evaluation tasks above. Because the released checkpoint also reflects SFT and model merging, we do not attribute every gain to RL alone. The overall pattern is nevertheless consistent with transfer of reusable behaviors—including planning, state tracking, tool selection, error recovery, and instruction adherence—across distinct agent environments.

\subsection{From Experts to a Single Model: Merging and Consolidation}
\label{subsec:merging}

The Marathon Expert is optimized for sustained execution over extended episodes, while the Sprint Expert preserves broader agentic capability. We combine them through the uniform parameter-space merge described in Section~\ref{sec:posttraining}, producing one checkpoint without inference-time routing or ensembling. Based on our experiments, we show that the model soup exceeds both experts on AutomationBench (29.0), while its WildClawBench score of 48.28 lies between the Marathon Expert (50.76) and the Sprint Expert (45.29). The soup is not the final model as we intend to further enhance the co-work ability, so we apply SAO to refine it on long-horizon interactions. Relative to the soup, Occamy-1.0 improves Claw-Eval from 79.2 to 82.2 and WildClawBench from 48.28 to 49.16. The final stage therefore shifts the balance toward co-work execution rather than improving every capability uniformly.


\subsection{Cost Efficiency and Reliability}
\label{sec:cost-efficiency}

\paragraph{Aggregate cost--performance.}
Cost efficiency is evaluated jointly with task performance rather than as a separate serving statistic. Figure~\ref{fig:aggregate-cost-performance} aggregates Claw-Eval, WildClawBench, AutomationBench, and GDPval under a common protocol. Occamy-1.0 lies near the low-cost knee of the observed Pareto frontier: compared with its Qwen3.6-35B-A3B starting checkpoint, it delivers a large increase in aggregate performance with only a modest change in per-task inference cost, while retaining a substantial cost advantage over the hosted frontier systems in this comparison.

\paragraph{Cost protocol.}
Performance results for larger frontier systems are obtained through hosted API evaluation. Compact models are evaluated with matched harness and decoding settings under local serving; each episode remains on one serving replica, and KV-cache capacity is provisioned so that no overflow or eviction is observed. Each stack is tuned to its best stable operating point. To obtain comparable monetary costs, we price measured token usage using the least expensive eligible OpenRouter provider available for each model at the time of analysis, applying the complete input, cache-read (when supported), and output price vector from that provider. All locally evaluated 35B-A3B checkpoints---Occamy-1.0, Qwen3.6-35B-A3B, Agents-A1, Nex-N2-mini, BigBang-1.0, and Ornith-1.5---use the same lowest complete 35B-A3B price vector. Holding unit prices fixed within this comparable-scale group prevents provider- or route-specific pricing differences from determining the cost ranking, so the comparison reflects differences in measured token usage under a common monetary basis. Appendix~\ref{app:eval_cost} reports the frozen price vectors and clarifies that this is a size-matched monetary proxy rather than a claim about checkpoint lineage or direct API availability. For model $m$ on benchmark $b$, we compute normalized performance $\tilde{s}_{m,b}=(s_{m,b}-\min_j s_{j,b})/(\max_j s_{j,b}-\min_j s_{j,b})$ and per-task cost $c_{m,b}=C_{m,b}/N_b$. Figure~\ref{fig:aggregate-cost-performance} plots $100\cdot\frac{1}{4}\sum_b\tilde{s}_{m,b}$ against $\frac{1}{4}\sum_b c_{m,b}$, giving all four benchmarks equal weight. This reports inference cost under a common pricing protocol, not complete deployment total cost of ownership. Serving details and per-benchmark frontiers are reported in Appendix~\ref{app:eval_cost}.

\paragraph{Execution efficiency and reliability.}
Table~\ref{tab:claweval-combined} compares Occamy-1.0 with the starting checkpoint on the combined Claw-Eval T/C tasks. In addition to success, we report token use, model turns, tool calls, wall-clock time, timeouts, invalid calls, and infrastructure failures under the same evaluation protocol.

Across the combined Claw-Eval T/C benchmark, Occamy consistently outperforms Qwen3.6-35B-A3B in success, efficiency, latency, and reliability. Occamy improves the Claw-Eval average score from 69.5 to 82.2 and Pass$^3$ from 54.8 to 71.4, while increasing execution success rate from 62.81\% to 77.55\%, a gain of 14.74 percentage points. It achieves these gains with 19.5\% fewer tokens, 1.2\% fewer model turns, and 15.2\% fewer tool calls. Occamy-1.0 also reduces trace wall time by 46.4\% and full-trial elapsed time by 36.6\%. It is substantially more reliable, reducing the timeout rate from 9.88\% to 2.18\%, the invalid-call rate from 1.85\% to 0.86\%, the tool-infrastructure failure rate from 1.14\% to 0.64\%, and the trial-infrastructure failure rate from 1.68\% to 0.50\%. Together, these measurements show that Occamy completes more tasks with less interaction, lower latency, and fewer execution failures under this protocol.

\begin{table}[ht]
\centering
\small
\setlength{\tabcolsep}{4pt}
\caption{Combined Claw-Eval T/C results from the updated runs. Claw-Eval Avg.\
and Pass$^3$ are strict metrics weighted across 161 T tasks and 38 C tasks;
trial success is measured over all 597 scheduled attempts per model. Token,
turn, tool-call, and trace-wall metrics are averaged over recorded trajectories.}
\label{tab:claweval-combined}
\resizebox{\textwidth}{!}{%
\begin{tabular}{lcccccc}
\toprule
\multicolumn{7}{c}{\emph{(a) Success and agent efficiency}} \\
\midrule
Model & Avg.\ Score & Pass$^3$ & Execution Success Rate
      & Tokens / Traj. & Turns / Traj. & Tool Calls / Traj. \\
\midrule
Qwen3.6-35B-A3B & 69.5 & 54.8 & 62.81\% & 185,999 & 7.28 & 14.24 \\
\textbf{Occamy-1.0} & \textbf{82.2} & \textbf{71.4} & \textbf{77.55\%}
      & \textbf{149,713} & \textbf{7.19} & \textbf{12.07} \\
\midrule
\multicolumn{7}{c}{\emph{(b) Latency and reliability}} \\
\midrule
Model & Trace Wall (s) & Trial Elapsed (s) & Timeout
      & Invalid Call & Tool Infra.\ Fail & Trial Infra.\ Fail \\
\midrule
Qwen3.6-35B-A3B & 74.58 & 222.44 & 9.88\% & 1.85\% & 1.14\% & 1.68\% \\
\textbf{Occamy-1.0} & \textbf{39.96} & \textbf{141.13} & \textbf{2.18\%}
      & \textbf{0.86\%} & \textbf{0.64\%} & \textbf{0.50\%} \\
\bottomrule
\end{tabular}%
}
\end{table}


\section{Analyses and Lessons Learned}
\label{sec:analysis}

\subsection{Approaches to Synthesizing Datasets with Learning Signals}

We draw three lessons from task construction. First, we select tasks with strong learning signals by retaining those where a stronger model outperforms a weaker one by a sufficient margin. During this process, score gaps caused by infrastructure instability are identified and filtered out. Second, environmental variation is a better predictor of training yield than nominal task difficulty. We characterize each task–environment instance by four components: workspace files, backend state, request wording, and grading criteria. Across task families, the rate of producing meaningful score gaps tracks how many of these components genuinely vary much more closely than the assigned difficulty level. This also reveals a limitation of template-based diversity: tasks generated from the same underlying structure may appear diverse through parameter changes while exposing the model to essentially the same problem. Third, tasks should provide partial credit rather than rely on binary success criteria. Partial credit better captures differences in evidence gathering and tool use, while binary objectives often collapse strong and weak models into the same outcome. We also do not treat trajectory length as informative by itself, since long trajectories may simply reflect repeated or failed tool interactions. In practice, tool-chain complexity and grader coverage are more reliable indicators of meaningful task difficulty.

\subsection{Lessons from SFT Data Ablations} 
 Our experiments suggest that SFT benefits most from trajectories with dense and reliable learning signals: correct outcomes, valid actions, and coherent intermediate steps. Data composition is equally important. Broad coverage across co-work, tool use, search, and other supporting capabilities produced more consistent gains than repeatedly scaling a narrow data source. Within the range studied, trajectory quality and capability coverage mattered more than sample count alone.\par For imperfect multi-turn trajectories, we apply hierarchical sanitization rather than a single filtering rule. We discard trajectories whose supervision is broadly unreliable, truncate trajectories when only the prefix remains valid, mask isolated assistant spans when the surrounding interaction is still useful, and repair or regenerate erroneous steps when a corrected continuation can be verified. 

\subsection{Lessons from Agentic Reinforcement Learning}

We revisit the final SAO run in Figure~\ref{fig:reward_explained_variance} to examine how verified reward and critic explained variance evolve during post-merge refinement. Across our RL runs, a recurring observation was transfer beyond the training distribution. When training the Marathon Expert using only its task pool, we also observed improvements on task-disjoint tool-calling and agentic benchmarks, without introducing data specific to those benchmarks. This pattern is consistent with the intended role of the RL stage and suggests that co-work training may strengthen reusable execution primitives that other agentic evaluations exercise in more isolated forms.

These results also illustrate why reward alone may not fully characterize the learning process. A co-work reward contract is necessarily an incomplete specification of good behavior. Graders certify task outcomes, whereas many properties that distinguish robust execution from fragile execution are not represented in the resulting scalar. Reward improvement remains an important signal, but a reward plateau does not necessarily imply an absence of learning. In one run, verified reward remained on an extended plateau while the policy's use of parallel tool calls varied substantially across steps. Reward began to improve only after this behavior became more stable. Viewed through the reward curve alone, the plateau resembles stagnation. Viewed together with the behavior trace, it is consistent with a reorganization phase in which the policy was changing how it invoked tools before those changes were reflected in task completion. Although this observation does not establish a causal sequence, it motivates monitoring policy behavior as a complementary training diagnostic. We therefore track tool-call structure, parallelism, interaction turns, and trajectory length alongside reward, and treat behavioral drift without reward movement as a signal that warrants inspection rather than automatic intervention.

Additionally, we also observed a characteristic early-stage learning pattern in SAO training. In our configuration, SAO typically requires a relatively long critic warmup before stable policy optimization begins. Even after this stage, the policy often exhibits substantial behavioral exploration while the reward remains nearly flat, indicating that meaningful policy changes may precede measurable reward improvement. One possible interpretation is that SAO accumulates behavioral diversity across successive policy versions, such that exploration unfolds over time rather than being concentrated within a single collection round. As a result, early reward curves alone may provide an incomplete picture of learning progress, and behavioral changes can serve as a useful complementary diagnostic during this phase.



\subsection{Reward Hacking and False Progress}

Reward hacking refers to behavior that obtains or preserves reward without satisfying the underlying task objective. For a cowork agent, this includes manipulating evaluator-visible state, accessing hidden evaluation data, weakening graders, fabricating evidence of completion, or exploiting superficial reward features. Ordinary errors or incomplete execution are not considered reward hacking unless they are used to influence evaluation.

\textbf{Methodology}. We audit complete trajectories rather than relying on reward or final-answer quality alone. The audit has three stages. First, deterministic detectors screen for environment manipulation, evaluator exploitation, representation manipulation, and reward-feature exploitation, supplemented by cross-rollout signals such as anomalously high reward with little work or unusual tool usage. Second, each trajectory is reviewed semantically in its task context, and all non-none cases receive an independent second review. Third, high-risk judgments are calibrated against the raw tool sequence and evaluator evidence. We confirm a case only when a specific action can be causally linked to evaluation manipulation; missing evidence remains unknown for human review. Confirmed cases are then used to update the first-stage detectors, forming an iterative audit loop.

\textbf{Observations}. Three lessons were particularly important. First, task feasibility is part of reward-hacking prevention: scaled synthetic tasks sometimes requested unavailable capabilities, such as image understanding through \texttt{OPENROUTER\_API\_KEY} without a usable API, allowing plausible hallucinations to be rewarded as false progress. Second, evaluation secrecy must cover both local and public channels. During training, agents could recover answers either from prematurely exposed grader artifacts or by locating the source dataset and reference answers on HuggingFace or other public repositories. We therefore inject grading artifacts only after the agent process is sealed and audit external retrieval for benchmark leakage. Third, RL data selection is itself an attack surface. Reward- or pass-rate-based filtering may promote exploitative trajectories as positive training examples. For example, an infeasible task requiring an invalid MCP tool may still receive non-zero \texttt{pass@16} if the policy bypasses the intended interface and reads the service's backing database. Once selected for RL, such trajectories turn an incidental exploit into a reinforced strategy. To prevent this, we apply the same reward-hacking guardrails used during RL training, including workspace isolation, tool-access restrictions, and execution-trace auditing, to the data-selection pipeline. In our latest audit of 4,352 trajectories, the four confirmed hacking rollouts all originated from repeated executions of the same underlying task. This concentration suggests that, at our model scale, observed reward hacking arose primarily from exploitable defects in task data and evaluation infrastructure, rather than from emergent, strategically deceptive behavior by the model itself.

\section{Conclusion}\label{sec:conclusion}	\textbf{Occamy-1.0 is a compact model built for co-work.}\ It is designed for long, stateful tasks that require an agent to coordinate tools, files, structured APIs, and productivity software over many steps. Rather than rebuilding general capability from a base model, we continue post-training from Qwen3.6-35B-A3B and concentrate learning on the coordination, recovery, and follow-through that real work demands.\par\smallskip	\textbf{The model is the product of an execution-centered post-training system.}\ Our data and environments connect task construction to runnable state transitions and task-level outcomes. The training infrastructure supports multiple harnesses while preserving token-exact trajectories, environment-state replay, and segment boundaries created by history rewrites. On top of this foundation, we train a Marathon Expert with SFT and HDPO, a Sprint Expert with SFT, combine their complementary behavior through model merging, and refine the merged checkpoint with SAO on a broad co-work mixture.\par\smallskip	\textbf{The results support co-work specialization at a practical model scale.}\ Across a broad suite of co-work benchmarks, Occamy-1.0 performs strongly among comparable-scale models and remains competitive with substantially larger systems on several long-horizon tasks. Its tool-calling, coding, and instruction-following results further show that the specialization does not reduce the model to a single benchmark profile. Important gaps remain in timeout robustness, native browser and visual interaction, and joint learning across subagents and multiple objectives. We release the model weights and a subset of the training data as a foundation for improving these capabilities and for studying practical co-work agents at smaller serving scales.
\clearpage
\section*{Contribution}
\addcontentsline{toc}{section}{Contribution}
\label{sec:contribution}

The title page uses collective authorship as \textbf{Accio Team}. This section records the complete contributor list.

\paragraph{Team members (alphabetical by last name).}
\begin{center}\small\renewcommand{\arraystretch}{1.15}\begin{tabular*}{\textwidth}{@{\extracolsep{\fill}}llll@{}} Wenhui Chen & Xinke Kong & Kunyu Shi & Qingcheng Zeng \\ Shiwen Cheng & Hongyu Li & Xiaojun Tang & Di Zhang \\ Hao Dong & Jiazheng Li & Bingquan Wang & Guannan Zhang \\ Chenda Duan & Junbo Li & Kesu Wang & Haochen Zhang \\ Ruixiang Feng & Qingchuan Li & Yuchen Wang & Tianlong Zhang \\ Zhong Guan & Yukun Lian & Sibo Wei & Tianyu Zhao\textsubscript{1} \\ Boqiang Guo & Chang Liu & Sicong Xie & Tianyu Zhao\textsubscript{2} \\ Xueyuan Han & Tianyu Liu & Xiaoying Xing & Yanjun Zheng \\ Haojie Hao & Zicheng Liu & Yi Xu & Jialong Zhu \\ Liangmeng Huang & Shuyi Ouyang & Zhijun Xu & Zijian Zou \\ Zhelong Huang & Yijun Pan & Hongwei Xue & \\ \end{tabular*}\end{center}


\clearpage
{
    \small
    \bibliographystyle{colm2024_conference}
    \bibliography{ref,benchmark_refs}

@article{officeqapro2026,
  title={OfficeQA Pro: An Enterprise Benchmark for End-to-End Grounded Reasoning},
  author={Opsahl-Ong, Krista and Singhvi, Arnav and Collins, Jasmine and Zhou, Ivan and Wang, Cindy and Baheti, Ashutosh and Oertell, Owen and Portes, Jacob and Havens, Sam and Elsen, Erich and Bendersky, Michael and Zaharia, Matei and Chen, Xing},
  journal={arXiv preprint arXiv:2603.08655},
  year={2026},
  url={https://arxiv.org/abs/2603.08655}
}

@misc{nexn22026,
  title        = {Nex-N2: An Agentic Model with Agentic Thinking},
  author       = {{Nex-AGI Team}},
  year         = {2026},
  howpublished = {GitHub repository},
  url          = {https://github.com/nex-agi/Nex-N2}
}

@misc{businessarena2026,
  title        = {Business Arena: Benchmarking LLM Agents in a Realistic Marketplace},
  author       = {Pan, Yijun and Lian, Yukun and Shi, Kunyu and Li, Junbo and Xue, Hongwei and Xie, Sicong and Zhang, Guannan and Xing, Xiaoying},
  year         = {2026},
  eprint       = {2608.08621},
  archivePrefix= {arXiv},
  primaryClass = {cs.AI},
  url          = {https://arxiv.org/abs/2608.08621}
}

@misc{realreplica2026,
  author       = {Lian, Yukun and Wei, Lei and Xie, Sicong and Zhang, Guannan and Wang, Kesu and Li, Hongyu and Jiang, Chenhao and Lin, Lanbo and Yang, Tianyuan and Guo, Xiaoyu and Cai, Li and Zhu, Jialong},
  title        = {Commerce Agent Bench: A Stateful Agent Benchmark for Long-Horizon Commerce and Business Workflows},  url          = {https://github.com/Accio-org/CommerceAgentBench},
  note         = {GitHub repository, v1.3.1},
  year         = {2026}
}

@article{wildclawbench2026,
  title={WildClawBench: A Benchmark for Real-World, Long-Horizon Agent Evaluation},
  author={Ding, Shuangrui and Dai, Xuanlang and Xing, Long and Ding, Shengyuan and Liu, Ziyu and JingYi, Yang and Yang, Penghui and Zhang, Zhixiong and Wei, Xilin and Fang, Xinyu and others},
  journal={arXiv preprint arXiv:2605.10912},
  year={2026},
  url={https://github.com/InternLM/WildClawBench}
}

@article{terminalbench2026,
  title={Terminal-Bench: Benchmarking Agents on Hard, Realistic Tasks in Command Line Interfaces},
  author={{Terminal-Bench Team}},
  journal={arXiv preprint arXiv:2601.11868},
  year={2026},
  url={https://www.tbench.ai/news/terminal-bench-2-1}
}

@misc{tau3bench2026,
  title={tau3-bench},
  author={{Sierra Research}},
  year={2026},
  howpublished={\url{https://taubench.com/}}
}

@article{automationbench2026,
  title={AutomationBench: Evaluating Intelligent Agents on Real-World Automation Workflows},
  author={{Zapier}},
  journal={arXiv preprint arXiv:2604.18934},
  year={2026},
  url={https://github.com/zapier/AutomationBench}
}

@misc{bfcl2026,
  title={Berkeley Function-Calling Leaderboard},
  author={{Berkeley Gorilla Team}},
  year={2026},
  howpublished={\url{https://gorilla.cs.berkeley.edu/leaderboard.html}}
}

@article{vitabench2025,
  title={VitaBench: Benchmarking Large Language Models for Tool-Augmented Agents in Real-World Interactive Scenarios},
  author={{VitaBench Team}},
  journal={arXiv preprint arXiv:2509.26490},
  year={2025},
  url={https://vitabench.github.io/}
}

@article{gdpval2025,
  title={GDPval: Evaluating AI Model Performance on Real-World Economically Valuable Tasks},
  author={{OpenAI}},
  journal={arXiv preprint arXiv:2510.04374},
  year={2025},
  url={https://openai.com/index/gdpval/}
}

@misc{claweval2026,
  title={Claw-Eval: Towards Trustworthy Evaluation of Autonomous Agents},
  author={Ye, Bowen and Li, Rang and Yang, Qibin and Liu, Yuanxin and Yao, Linli and Lv, Hanglong and Xie, Zhihui and An, Chenxin and Li, Lei and Kong, Lingpeng and Liu, Qi and Sui, Zhifang and Yang, Tong},
  year={2026},
  eprint={2604.06132},
  archivePrefix={arXiv},
  primaryClass={cs.AI},
  url={https://arxiv.org/abs/2604.06132}
}

@misc{hou2026singlerolloutasynchronousoptimizationagentic,
      title={Single-Rollout Asynchronous Optimization for Agentic Reinforcement Learning}, 
      author={Zhenyu Hou and Yujiang Li and Jie Tang and Yuxiao Dong},
      year={2026},
      eprint={2607.07508},
      archivePrefix={arXiv},
      primaryClass={cs.LG},
      url={https://arxiv.org/abs/2607.07508}, 
}

@misc{li2026compactionrlreinforcementlearningcontext,
      title={CompactionRL: Reinforcement Learning with Context Compaction for Long-Horizon Agents}, 
      author={Yujiang Li and Zhenyu Hou and Yi Jing and Jie Tang and Yuxiao Dong},
      year={2026},
      eprint={2607.05378},
      archivePrefix={arXiv},
      primaryClass={cs.LG},
      url={https://arxiv.org/abs/2607.05378}, 
}

@misc{evalscope2024,
  title={{EvalScope}: Evaluation Framework for Large Models},
  author={ModelScope Team},
  year={2024},
  url={https://github.com/modelscope/evalscope}
}

@misc{yue2025vapoefficientreliablereinforcement,
      title={VAPO: Efficient and Reliable Reinforcement Learning for Advanced Reasoning Tasks}, 
      author={Yu Yue and Yufeng Yuan and Qiying Yu and Xiaochen Zuo and Ruofei Zhu and Wenyuan Xu and Jiaze Chen and Chengyi Wang and TianTian Fan and Zhengyin Du and Xiangpeng Wei and Xiangyu Yu and Gaohong Liu and Juncai Liu and Lingjun Liu and Haibin Lin and Zhiqi Lin and Bole Ma and Chi Zhang and Mofan Zhang and Wang Zhang and Hang Zhu and Ru Zhang and Xin Liu and Mingxuan Wang and Yonghui Wu and Lin Yan},
      year={2025},
      eprint={2504.05118},
      archivePrefix={arXiv},
      primaryClass={cs.AI},
      url={https://arxiv.org/abs/2504.05118}, 
}

@misc{digitous2023llmslerp,
  author  = {{Digitous} and {CalderaAI}},
  title   = {{LLM-SLERP-Merge}: Spherical Linear Interpolation for Merging Language Models},
  year    = {2023},
  url     = {https://github.com/Digitous/LLM-SLERP-Merge},
  urldate = {2026-08-26},
  note    = {GitHub repository}
}

@misc{openai2026gpt56syscard,
  title        = {{GPT-5.6} System Card},
  author       = {{OpenAI}},
  year         = {2026},
  month        = jul,
  howpublished = {\url{https://deploymentsafety.openai.com/gpt-5-6}},
  note         = {Accessed: 2026-08-26}
}

@article{glm2026glm5,
  title   = {{GLM-5} Technical Report},
  author  = {{Z.ai} and {GLM Team}},
  journal = {arXiv preprint arXiv:2602.15763},
  year    = {2026}
}

@misc{qwen2026qwen38max,
  title        = {{Qwen3.8-Max}},
  author       = {{Qwen Team, Alibaba Cloud}},
  year         = {2026},
  month        = aug,
  howpublished = {\url{https://qwenlm.github.io/blog/qwen3.8/}},
  note         = {API model ID: \texttt{qwen3.8-max}. Accessed: 2026-08-26}
}

@article{deepseekai2026v4,
  title   = {{DeepSeek-V4}: Towards Highly Efficient Million-Token Context Intelligence},
  author  = {{DeepSeek-AI}},
  journal = {arXiv preprint arXiv:2606.19348},
  year    = {2026}
}

@article{bai2026agentsa1,
  title   = {Scaling the Horizon, Not the Parameters: Reaching
             Trillion-Parameter Performance with a 35B Agent},
  author  = {Bai, Lei and Cao, Zongsheng and Chen, Yang and others},
  journal = {arXiv preprint arXiv:2606.30616},
  year    = {2026}
}

@article{bigbang2026,
  title   = {{BigBang}: Pursuing Open-Ended Intelligence through
             Self-Evolving Synthesis of Verifiable Frontier Tasks},
  author  = {{The BigBang Team}},
  journal = {Preprint. Technical Report},
  year    = {2026},
  note    = {\url{https://endlessfrontier.tech/assets/paper.pdf}}
}

@misc{ornith2026,
  title        = {{Ornith-1.5}},
  author       = {{DeepReinforce}},
  year         = {2026},
  month        = aug,
  howpublished = {\url{https://huggingface.co/ornith-ai/Ornith-1.5-35B-A3B}},
  note         = {MIT license. Accessed: 2026-08-26}
}

@article{yan2026act,
  title={Act wisely: Cultivating meta-cognitive tool use in agentic multimodal models},
  author={Yan, Shilin and Tong, Jintao and Xue, Hongwei and Tang, Xiaojun and Wang, Yangyang and Shi, Kunyu and Zhang, Guannan and Li, Ruixuan and Zou, Yixiong},
  journal={arXiv preprint arXiv:2604.08545},
  year={2026}
}

@misc{qwen2026qwen36_35b_a3b,
  title        = {{Qwen3.6-35B-A3B}},
  author       = {{Qwen Team}},
  year         = {2026},
  howpublished = {Hugging Face model card},
  url          = {https://huggingface.co/Qwen/Qwen3.6-35B-A3B},
  note         = {Accessed: 2026-08-27}
}

@inproceedings{wortsman2022modelsoups,
  title     = {Model Soups: Averaging Weights of Multiple Fine-Tuned Models Improves Accuracy without Increasing Inference Time},
  author    = {Wortsman, Mitchell and Ilharco, Gabriel and Gadre, Samir Yitzhak and Roelofs, Rebecca and Gontijo-Lopes, Raphael and Morcos, Ari S. and Namkoong, Hongseok and Farhadi, Ali and Carmon, Yair and Kornblith, Simon and Schmidt, Ludwig},
  booktitle = {Proceedings of the 39th International Conference on Machine Learning},
  year      = {2022}
}

@inproceedings{yadav2023tiesmerging,
  title     = {{TIES-Merging}: Resolving Interference When Merging Models},
  author    = {Yadav, Prateek and Tam, Derek and Choshen, Leshem and Raffel, Colin and Bansal, Mohit},
  booktitle = {Advances in Neural Information Processing Systems},
  year      = {2023},
  eprint    = {2306.01708},
  archivePrefix = {arXiv},
  primaryClass  = {cs.LG},
  url       = {https://arxiv.org/abs/2306.01708}
}

@article{zhou2023instruction,
  title={Instruction-following evaluation for large language models},
  author={Zhou, Jeffrey and Lu, Tianjian and Mishra, Swaroop and Brahma, Siddhartha and Basu, Sujoy and Luan, Yi and Zhou, Denny and Hou, Le},
  journal={arXiv preprint arXiv:2311.07911},
  year={2023}
}

@misc{openclaw,
  title        = {{OpenClaw}: Your Assistant, on Your Devices, in Your Chats},
  author       = {{OpenClaw Foundation}},
  year         = {2026},
  howpublished = {GitHub repository},
  url          = {https://github.com/openclaw/openclaw},
}

@misc{hermes,
  title        = {{Hermes Agent}: The Agent That Grows With You},
  author       = {{Nous Research}},
  year         = {2026},
  howpublished = {GitHub repository},
  url          = {https://github.com/NousResearch/hermes-agent}
}

@misc{dressage_github,
  author       = {Liangmeng Huang and Qingchuan Li and Hongwei Xue and Shilin Yan and {Dressage Contributors}},
  title        = {{Dressage}: Scalable {RL} for Any Agent and Any Sandbox},
  year         = {2026},
  howpublished = {\url{https://github.com/Accio-Lab/Dressage}}
}
}

\clearpage
\appendix
\section*{Appendix}
\addcontentsline{toc}{section}{Appendix}

\section{Data Construction and Admission Details}
\label{app:data_details}

\subsection{Task Design Principles}
\label{app:task_design_principles}

Although the environment-first and capability-first synthesis routes differ in
their starting points, they are governed by the same six design principles.
We state these principles in full below; Section~\ref{subsec:task_validity}
provides a concise overview. Throughout this appendix, we use the
task--environment notation $(Q_i, W_i, T_i, G_i)$ introduced in
Section~\ref{subsec:task_environment_spec}.

\paragraph{Professional grounding and substantive difficulty.}
This principle requires each task to be framed as a plausible professional
assignment with a concrete objective, a clear deliverable, a downstream use
when relevant, and constraints that directly affect success. Its files, service
state, and tool interfaces form a coherent work setting. Difficulty arises from
substantive demands such as evidence reconciliation, conflicts in authority or
timing, multi-step execution, and recovery from failures. Accordingly, we
exclude artificial sources of difficulty, such as ambiguous instructions,
unnecessarily long prompts, and arbitrary requirements unrelated to the task
objective.

\paragraph{Candidate-visible feasibility.}
A task must be solvable under the same information and action constraints faced
by the evaluated policy. Every referenced path, entity, and record must exist,
and the evidence needed for completion must be reachable through the visible
workspace and permitted tools. Any conflicts introduced during task construction, such as conflicts in source
authority or timing, must be resolvable using only evidence visible to the
evaluated policy. Finally, the permitted tools and services must support at least one valid
completion path.

\paragraph{Faithful environment semantics.}
The environment may maintain private service state in addition to the visible
workspace to model realistic and persistent service behavior. The policy
interacts with the service through declared interfaces, which expose only the
observations and actions permitted by the task. The contract for each interface
must faithfully describe the observations and state transitions implemented by
its backend, including permission checks, failure modes, retry semantics, and
repeated-call behavior. The system records each invocation and its realized
effects, and persists the resulting state for verification.

\paragraph{Evidence-grounded grading.}
Grading is applied to sealed execution evidence, including delivered artifacts,
final service state, and recorded actions; an agent's completion claim is not
sufficient evidence. Deterministic checks target task-relevant entities,
fields, relations, and state transitions, and bounded semantic judgment is used
when correctness cannot be reduced to those checks. A valid grader accepts substantively correct execution paths while rejecting
degenerate strategies such as no-ops, constant outputs, unauthorized side
effects, and other shortcuts that satisfy superficial checks without completing
the requested work.

\paragraph{Information isolation and precommitment.}
From the task package, the evaluated policy receives only the public request,
the agent-visible portion of the initial world state, and the public tool
contracts. Private fixtures, backend routes, construction evidence,
reference-execution outputs, and the private completion and grading contract
remain outside the policy's view. The grading contract is frozen before
evaluated rollouts begin and applied to sealed evidence without being revised
in response to policy outputs.

\paragraph{End-to-end alignment.}
The components of a task contract must be designed and validated jointly. Any state-dependent public requirement must be grounded in the initial
environment. Each required observation or action must have a
feasible path through the permitted tools, and every public requirement must
map to a corresponding success condition in the grading contract. Conversely,
every graded condition must be justified by the public request and verifiable
from evidence available within the evaluated policy's information boundary.
\subsection{Capability and Environment Structures}
\label{app:capability_environment_structures}
Following these design principles, we organize task construction around two
complementary structures: a capability structure and an environment structure.
Together, they define the joint capability--environment design space used by
both synthesis routes.
\paragraph{Capability structure.}
For each reference task, this analysis produces a dependency structure over
the capabilities required to complete it. Its nodes capture task-relevant
operations, including source discovery and identity resolution, information
extraction and reconciliation, reasoning under exceptions and uncertainty,
evidence-grounded decision making, cross-tool execution, artifact construction,
and outcome verification. Edges encode prerequisite or load-bearing
relationships between these capabilities. The resulting dependencies may form
a sequence, branch across independent evidence sources, or converge as multiple
analyses feed into a single decision or deliverable.

The structure guides task construction by recording the substantive
dependencies that define the task's difficulty while leaving the concrete
execution path open. A policy may therefore use different plans, tools, and
intermediate representations, provided that the resulting execution satisfies
the task contract. Separating the capability structure from its concrete
realization allows us to distinguish variation in presentation or environment
from changes in the capability composition itself.

\paragraph{Environment structure.}
An environment structure describes how task evidence and mutable state are
organized, how the agent can access that evidence, and which state transitions
it can perform. For construction and coverage control, it captures the
workspace organization and source types, the topology of tools and services,
mutable entities and permitted transitions, and the deliverable or target-state
family. A given environment structure can support multiple capability
compositions, while the same capability structure can be realized across
different environments.

\subsection{Task Variation and Collection-level Coverage}
\label{app:task_variation_coverage}

After each round of task construction, we analyze the coverage of the resulting
collection. We examine the professional context and expected output of each
task, the capabilities required and the dependencies among them, and the
environment and tools through which the task is executed. We also consider
difficulty factors such as conflicting evidence, uncertainty, tool failures, and
recovery. This analysis ensures that surface-level variants of the same
workflow are not counted as additional structural diversity. It also shows
which capabilities, environments, and workflow structures remain
underrepresented. We use these findings to guide the next round of
environment-first and capability-first synthesis.

\subsection{Environment-first Task Synthesis}
\label{subsec:environment_first_synthesis}

Environment-first synthesis starts from a digital environment that can support
and verify a useful form of work. We organize this process into four recurring
construction patterns: evidence-grounded investigation and decision making,
workspace artifacts and code repair, cross-service state workflows, and
package-grounded integration. Although they draw on different resources and
interfaces, all four follow the same executable loop: we select a coherent set
of interfaces and assets, construct the request and initial state around them,
materialize the environment, and run a reference execution using the same
interfaces available during rollout. The specific construction procedure for
each pattern is described below.

\paragraph{Evidence-grounded investigation and decision.}
These tasks require the agent to gather and reconcile evidence from multiple
sources to produce an analysis or structured decision. We assemble connected
documents, workspace assets, and service interfaces whose records can be joined
within a single investigation, for example by customer, account, event, or time
period. We jointly construct the request, initial state, target conclusion, and
grading criteria. Difficulty arises from conflicting sources, authority rules,
temporal cutoffs, look-alike entities, or exceptions that change the apparent
conclusion. A reference execution confirms that every load-bearing fact is
available through the same sources and interfaces used during rollout. The
grader checks the resulting analysis or decision, while negative executions
cover missing evidence, unauthorized state changes, and entity mismatches.

\paragraph{Workspace artifacts and code repair.}
These tasks center on creating, transforming, or repairing files in a
workspace. For document-grounded tasks, we retrieve complete and parseable
files from maintained public datasets and GitHub repositories. When these
sources do not provide enough context for a coherent professional task, we add
a small task-specific record, such as a current policy, an exception notice, or
a prior draft. After assembling the workspace, we analyze the files to identify a professional
objective and deliverable that require substantive use of their content. We
then formulate the request around this objective and verify that the required
result is derivable from the files available to the agent.

For structured artifacts such as schedules, reconciliation reports, and
deployment configurations, deterministic generators create consistent inputs
and expected outputs. Code-repair tasks instead select the target module and
its relevant callers or helpers from a single repository, then inject two to
four independent semantic faults. The prompt describes the observable symptoms
without revealing the faulty locations. Non-code artifacts are checked against
source-backed criteria or deterministic expected outputs, while code repairs
are validated using per-fault reproduction tests and baseline regression tests
on both the faulty and repaired versions.

\paragraph{Cross-service state workflows.}
These tasks require the agent to coordinate reads and persistent state changes
across multiple services. We select services linked by shared entities or
workflow dependencies, then populate them with records, rules, and consistent
identifiers. We specify the required final state together with any ordering and
side-effect constraints.

Operations across services may depend on one another. For example, the agent may need to retrieve a customer or order identifier
before updating a downstream record, check capacity or policy before performing
a write, or check whether a failed call has already changed the service state
before retrying it. A reference execution
completes the workflow through the same public tools available during rollout.
The grader checks the final service states, recorded calls, and mutation logs.
We also test executions that omit required updates, modify the wrong records,
duplicate a write, or perform unauthorized actions.

\paragraph{Package-grounded integration.}
These tasks require the agent to build a small application or integration
using an existing software package. We retrieve the package's complete source
tree from a maintained collection of public GitHub repositories, preserve its
directory structure, and verify that it can be installed and imported. We then
define an application goal, prepare a runnable but incomplete starter, and
create visible inputs, held-out test cases, and a reference implementation.
Difficulty comes from using the package's actual APIs correctly while
satisfying task-specific input, output, and error-handling requirements.

The reference implementation is run in an isolated environment to establish
the expected behavior. Runtime tracing confirms that candidate solutions use
the selected package and task inputs. We also vary the test inputs and require
corresponding changes in the output, so that hard-coded or package-bypassing
solutions fail.

\paragraph{Shared generation gates.}
We retain a task only if all prompt-referenced assets exist, the agent-visible
state contains the evidence needed to solve it, and private fixtures or target
outputs remain outside that view. The public tool schemas must agree with the
running services, the reference execution must succeed from the provisioned initial
state, and negative executions must receive appropriately lower grades. A final
alignment review covers the request, workspace, service behavior, intended
execution, and grader. Tasks that pass these checks enter the
shared canonicalization and training-admission procedure.

\subsection{Capability-first Task Synthesis}
\label{subsec:capability_first_synthesis}

Capability-first synthesis begins by selecting a capability structure from
Section~\ref{app:capability_environment_structures} as the target for a new
task, before choosing its specific files, facts, or setting. The selected
capabilities and their dependencies determine what supporting evidence we
seek. We then retrieve and verify independent source materials. The evidence
they provide determines the concrete entities, facts, and scope of the
resulting task. We author the public request only after this evidence base has
been established.

\paragraph{Capability target freeze.}
We derive target capability structures from a collection of real-world
reference tasks. For each task, we identify its professional context, the roles
of its source materials, the required deliverable, and the factors contributing
to its difficulty. We also compare successful and unsuccessful executions of
the same task to identify the capabilities and dependencies that are important
for successful completion.

An independent review checks that the resulting structure generalizes beyond
task-specific content, preserves the essential capability dependencies, and
yields evidence requirements that can be met using discoverable and verifiable
sources. After revision and structural validation, we freeze the target.
Subsequent stages receive only this de-identified capability specification; the
original task text, instance-specific answer, grading rubric, and raw execution
trajectories are withheld.

\paragraph{Source requirements and evidence closure.}
For each required capability, we specify the source content and constraints
needed to support it, such as relevant entities, relationships, authority, time
period, and file format. For example, a financial task for reconciliation capability may require a bank statement, an internal transaction ledger, and an adjustment policy covering the same accounts and reporting period. We distinguish facts that must be grounded in sources
from task-level rules, such as the audience, cutoff date, and deliverable
format, that may be defined in the public request.

We retrieve candidate sources from an existing semantic catalog and through
targeted search over provenance-tracked collections. The files are screened for
provenance, integrity, parseability, duplication, and overlap, then analyzed
for the required evidence under independent review. We repeat targeted search
for any remaining gaps. If the complete capability structure cannot be
supported, we retain only the largest useful subset supported by verified
sources and construct a smaller task rather than inventing missing facts.

\paragraph{Source-grounded task authoring.}
Once the required evidence has been assembled, task authoring combines the
frozen capability target with the verified source files and a professional
scenario. The scenario specifies the agent's role, the event motivating the
work, the intended audience, the downstream use, and the required deliverable.
Together, these elements determine the work context, the role of each source,
and the public request.

An independent review checks that the task requires substantive use of the
provided files, is feasible from candidate-visible evidence, produces a useful
professional deliverable, and can be graded fairly. Difficulty arises from
relationships within and across the sources, such as conflicts in authority,
differences in version or reporting period, entity joins, exceptions, and
missing values. The public request may introduce task-level constraints, but it
may not introduce unsupported facts about the source materials. After revision
and validation, we freeze the request and place the verified source files in
the agent-visible workspace. Capability analyses and other construction records
remain private.

\paragraph{Blind reference execution.}
We next ask a strong reference solver to complete the frozen task in an isolated
sandbox. It receives the same public request, workspace files, and tools that
will be available to an evaluated policy, with no access to private source
analysis, grading criteria, or privileged hints. The run preserves the complete
trajectory, tool interactions, intermediate work, and submitted artifacts. It
establishes an executable reference path, exposes residual ambiguity in the
request or sources, and produces an artifact that can be parsed during grader
construction. We treat this artifact as an auditable completion hypothesis, and
the public request as well as source-verifiable evidence remain authoritative. The
reference solver's incidental choices of layout, wording, or implementation do
not become unique requirements in our real implementation.

\paragraph{Evaluation contract construction.}
After completing the reference execution, the reference solver uses the frozen
public request together with private construction evidence, including the
verified source analyses, reference artifact, and execution record, to draft a
task-specific evaluation contract. The contract maps each public requirement
to an independently checkable criterion, identifies the supporting evidence,
and specifies full- and partial-credit conditions, acceptable alternatives,
and material failures.

An independent review checks that all public requirements are covered, every
source-dependent criterion is grounded in the provided evidence, and choices
specific to the reference execution have not been turned into mandatory
requirements. After any necessary revision, the contract is frozen before
evaluated rollouts begin. The task, reference evidence, and frozen evaluation
contract then enter the shared canonicalization and training-admission pipeline.

\subsection{Canonicalization, Verification, and Training Admission}
\label{app:canonicalization_admission}
The procedure operates at both the task and episode levels. Task-level
validation establishes that a task--environment package is executable and
gradable. Episode-level admission then checks the execution evidence and
outcome of each run before converting eligible episodes into SFT records or RL
trajectories.

\paragraph{Canonical identity and provenance.}
Every task release binds the public request, initial state, public tool schemas,
realized handler implementation and build identity, execution protocol,
immutable runtime or container identities, grader, construction lineage, and
split assignment by content hash. Ephemeral endpoint addresses are recorded in
execution receipts rather than treated as task identity. Authoring and trajectory teaching are recorded as different
roles.  Held-out evaluation prompts, worlds, private grading material, and raw
evaluation trajectories are excluded from both roles.  A task enters a
training collection only when its release explicitly authorizes that use; a
successfully validated but evaluation-only or diagnostic package remains
ineligible.  Because materialization can change the realized services and
handlers, cross-release identity checks are repeated after provisioning over
the task, prompt, initial state, realized contract, service set, grader, and
construction lineage.  Authoring-stage identifiers alone are not treated as
sufficient evidence of uniqueness.

\paragraph{Executable and semantic gates.}
Admission requires static schema and reference checks, separate recording of
requested and realized tool handlers, rejection of unsupported or semantically
invalid substitutions, a successful reference execution through the public interfaces,
negative validation executions that receive lower grades, and a final
cross-component review of the request, world, tools, and grader.  We deduplicate
using content and realized-runtime identities rather than ordinal identifiers
or truncated task names.  Contamination checks cover prompts, world state,
tool and service interfaces, graders, and source lineage.  Infrastructure
failures, incomplete finalization, ambiguous outcomes and grades, and post-release mutations
are quarantined rather than converted into policy failures.

\paragraph{Reachability, discoverability, and service-side evidence.}
A reference execution establishes that a valid execution path exists; it does not
establish that a policy can discover that path.  When a declared step is pinned to an
exact \texttt{(method, URL)} pair, every object identifier embedded in that
route must also be obtainable from policy-visible world evidence, the public
request, or an enumerable search surface.  We therefore evaluate
discoverability as a separate admission gate from executable feasibility.  For
state-mutating service tasks admitted under this pipeline, the service-side
append-only audit ledger is reconciled bidirectionally with the harness
tool-dispatch log: every dispatched mutation must have a matching service
record, and every service-side mutation must originate from a recorded
dispatch.  Missing identifiers, unresolved routes, or incomplete ledger
evidence are treated as data or infrastructure failures rather than as model
failures.

\paragraph{Trajectory and reward admission.}
Teacher demonstrations must preserve the complete multi-turn interaction,
including structured tool calls, observations, side effects, and the final
verified state.  RL records additionally bind the exact model-side token stream,
rollout log-probabilities, response loss mask, policy version, termination
reason, and task-level reward.  The reward used by optimization is named and
versioned: an official frozen outcome score is not interchangeable with an
available-score or partial-credit diagnostic.  Clean behavioral failures may
be retained when the task contract calls for them, whereas sandbox, gateway,
environment, and grader failures never become zero-reward policy samples.

\paragraph{Release sealing.}
Each admitted release is create-only and contains a manifest of inputs,
versions, counts, byte sizes, and hashes; per-stage admission statistics; a
train/validation split fixed before training; and a marker-last completion
receipt.  Any correction creates a new release identity.  This makes the data
mixture reproducible without relying on mutable directory names or informal
checkpoint labels.

\subsection{Data Mixture and Coverage Control}
\label{app:data_mixture_coverage}

The training mixture has three layers with different units and admission
criteria: executable task--environment packages, verified SFT demonstrations,
and on-policy RL trajectories.  We report them separately rather than using a
single ``number of examples,'' since one task can produce multiple episodes and
one long episode can contain several loss-bearing segments. Table~\ref{tab:data-artifact-classes} lists the unit reported for each class and the release statistics we require before it enters training.

\begin{table}[t]
\centering
\small
\caption{Data artifacts are counted using the unit that enters the corresponding stage. Sealed internal manifests retain release counts and token totals.}
\label{tab:data-artifact-classes}
\begin{tabular}{p{0.20\linewidth}p{0.25\linewidth}p{0.43\linewidth}}
\toprule
Artifact class & Unit reported & Required release statistics \\
\midrule
Task and environment & executable task--environment package & tasks, environments, realized services and handlers, reference/negative executions, training authorization \\
SFT demonstration & verified episode and trainable tokens & source release, author and teacher roles, domain, turns, tool calls, verification result, rendered tokens \\
RL trajectory & episode, segment, and trainable tokens & checkpoint and policy version, reward specification, termination, tokens/log-probabilities/masks, infrastructure disposition \\
\bottomrule
\end{tabular}
\end{table}

The mixture covers end-to-end co-work together with supporting tool-calling,
coding, and search data.  Coverage is measured over domains, environment types,
realized tools and services, dependency topology, deliverable type, grader
family, turns, trajectory length, recovery events, and trainable-token share.
Requested tool diversity is reported separately from realized runtime coverage,
so handler substitution cannot inflate the apparent breadth of the collection.

Student pass@\(k\), trajectory length, and tool-chain depth are useful difficulty
signals, but none is the sole admission rule.  A hard-only mixture can collapse
the reward signal or over-represent one environment shape, while an easy-only
mixture may fail to improve planning and recovery.  We therefore apply
capability and environment coverage floors, token-share caps, and stability
requirements before using difficulty to prioritize examples.  These quantities are tracked in sealed internal manifests. This report publishes the deduplicated SFT union in the main SFT data table while keeping task-package, SFT, and RL accounting distinct.

\section{Evaluation Protocol and Benchmark Registry}
\label{app:evaluation_protocols}

\subsection{Common Serving and Reporting Protocol}

All checkpoints reproduced by us are served identically: SGLang with the updated setting (\texttt{mem-fraction-static}$=0.85$, 96 in-flight requests per replica, 384
concurrent), with a $262{,}144$-token context in which prompt and generation
share a single budget, so that all cross-benchmark variation comes from the
harnesses rather than from the serving stack. Each benchmark is then pinned to
its own upstream harness and run at that harness's native protocol.

\subsection{Per-benchmark Cost--Performance Frontiers}
\label{app:eval_cost}

\begin{figure}[H]
\centering
\includegraphics[width=\linewidth]{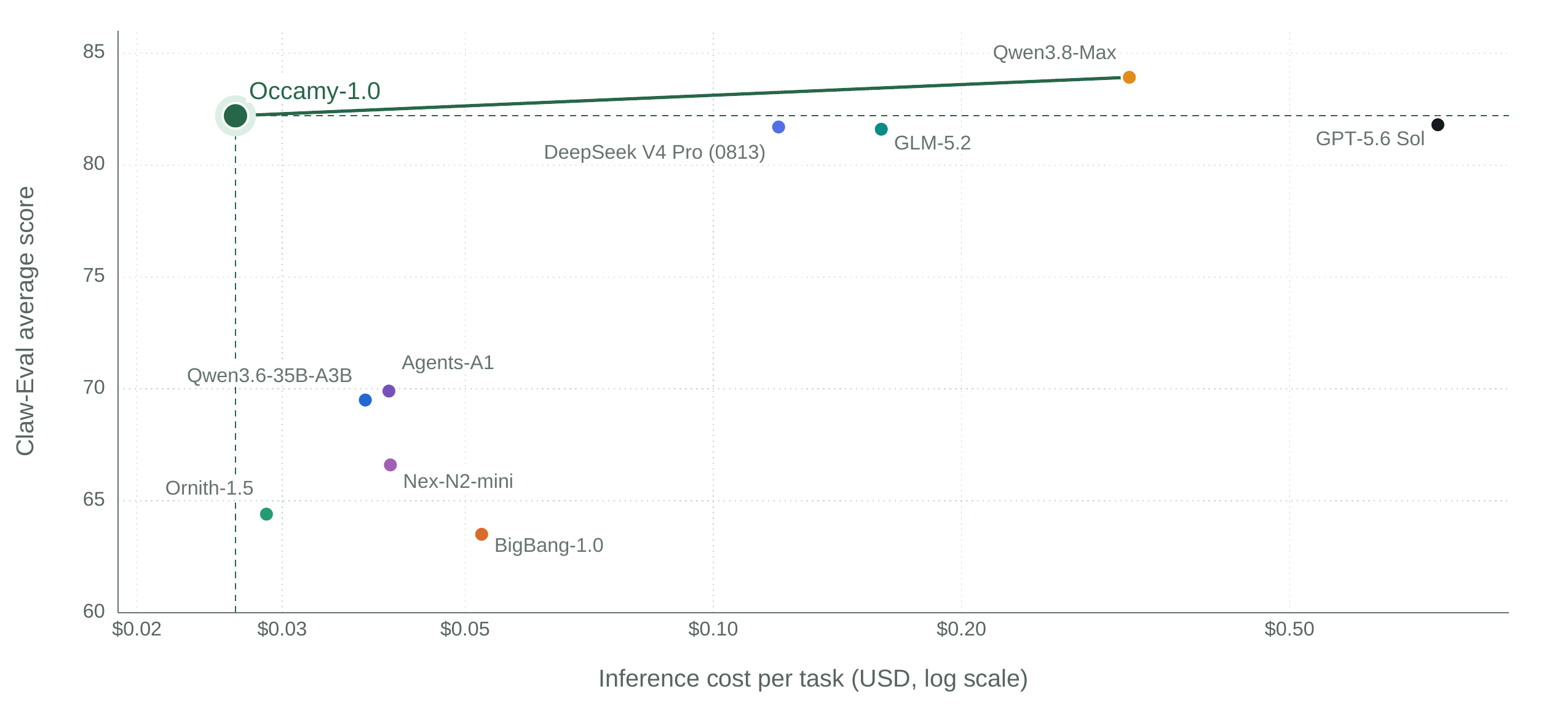}
\caption{Claw-Eval cost--performance frontier (199 tasks).}
\label{fig:cost-claweval}
\end{figure}

\begin{figure}[H]
\centering
\includegraphics[width=\linewidth]{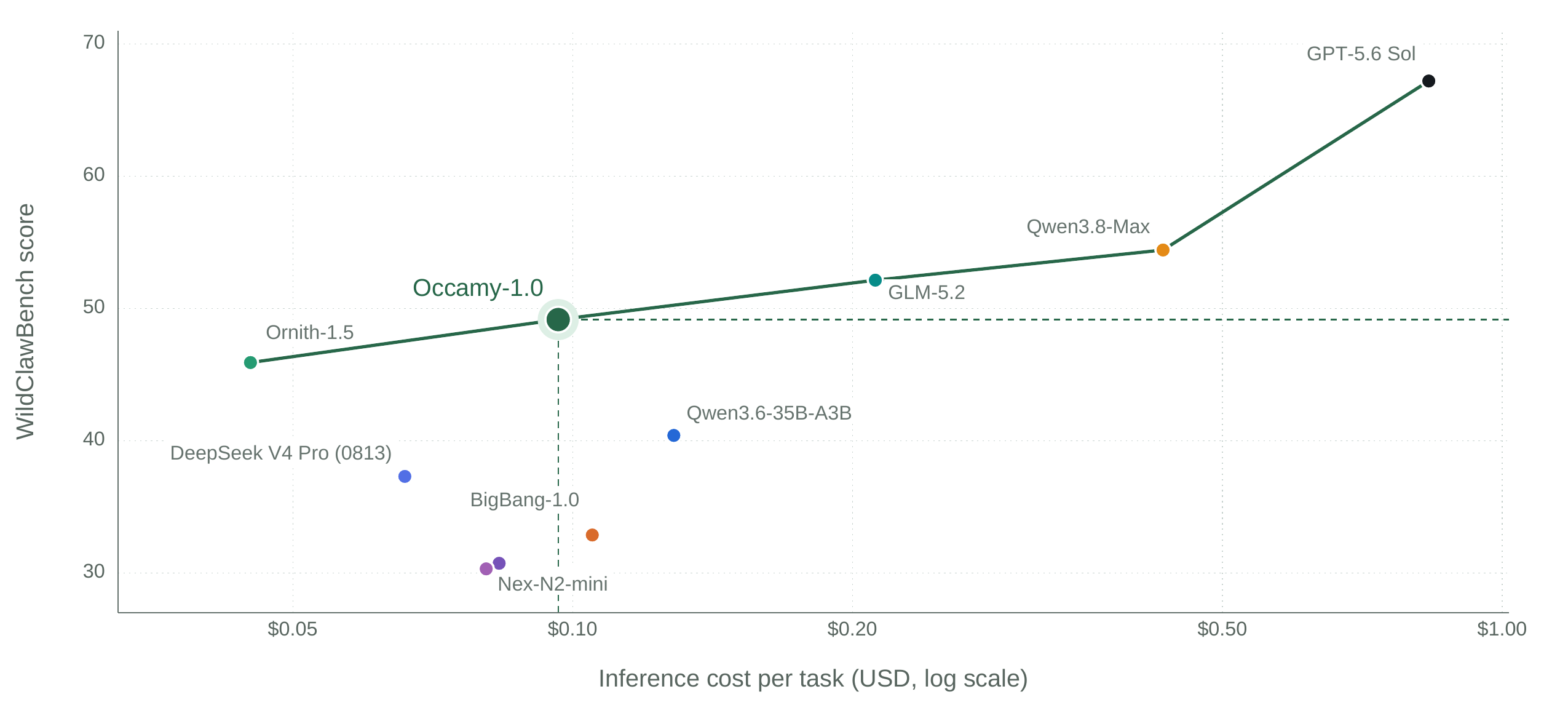}
\caption{WildClawBench cost--performance frontier (60 tasks).}
\label{fig:cost-wildclawbench}
\end{figure}

\begin{figure}[H]
\centering
\includegraphics[width=\linewidth]{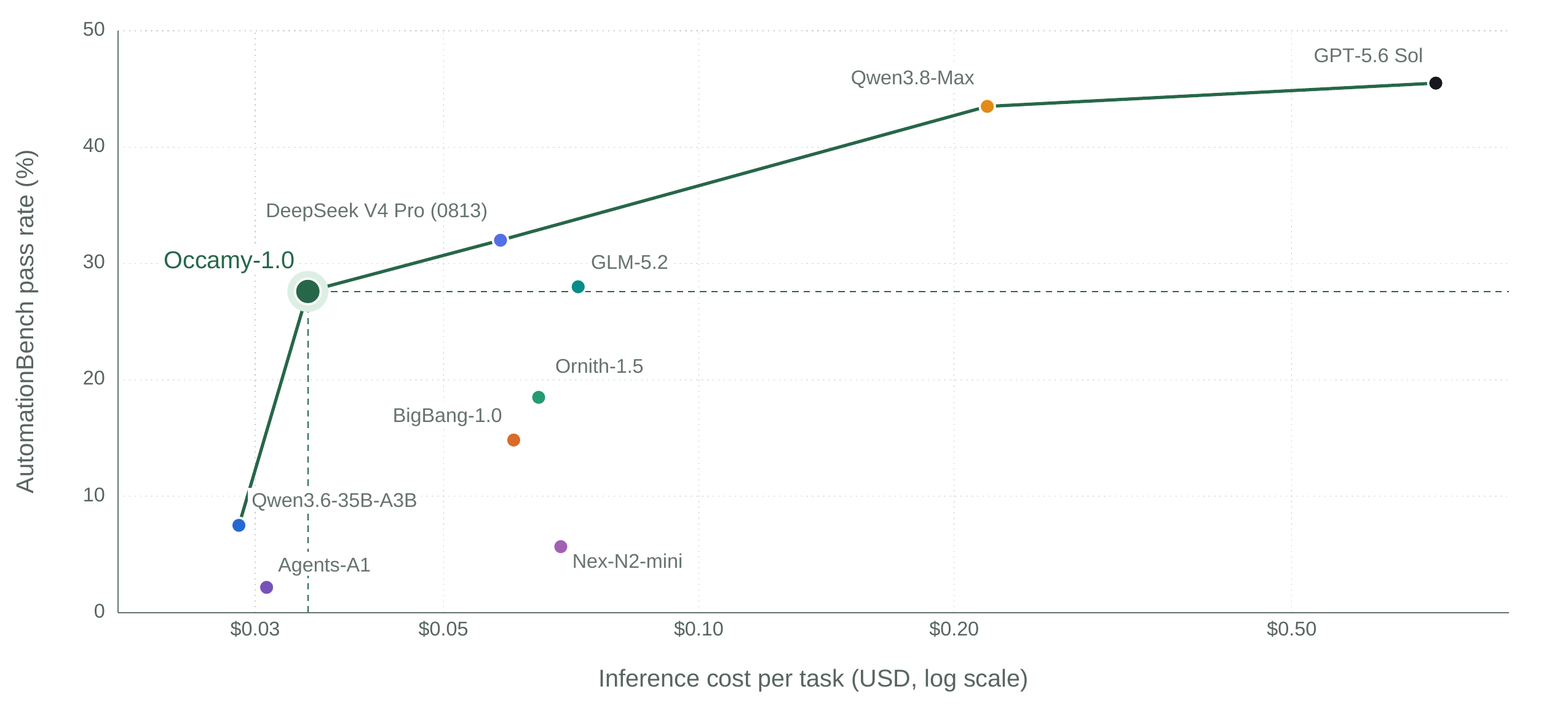}
\caption{AutomationBench strict pass-rate cost--performance frontier (600 tasks).}
\label{fig:cost-automationbench}
\end{figure}

\begin{figure}[H]
\centering
\includegraphics[width=\linewidth]{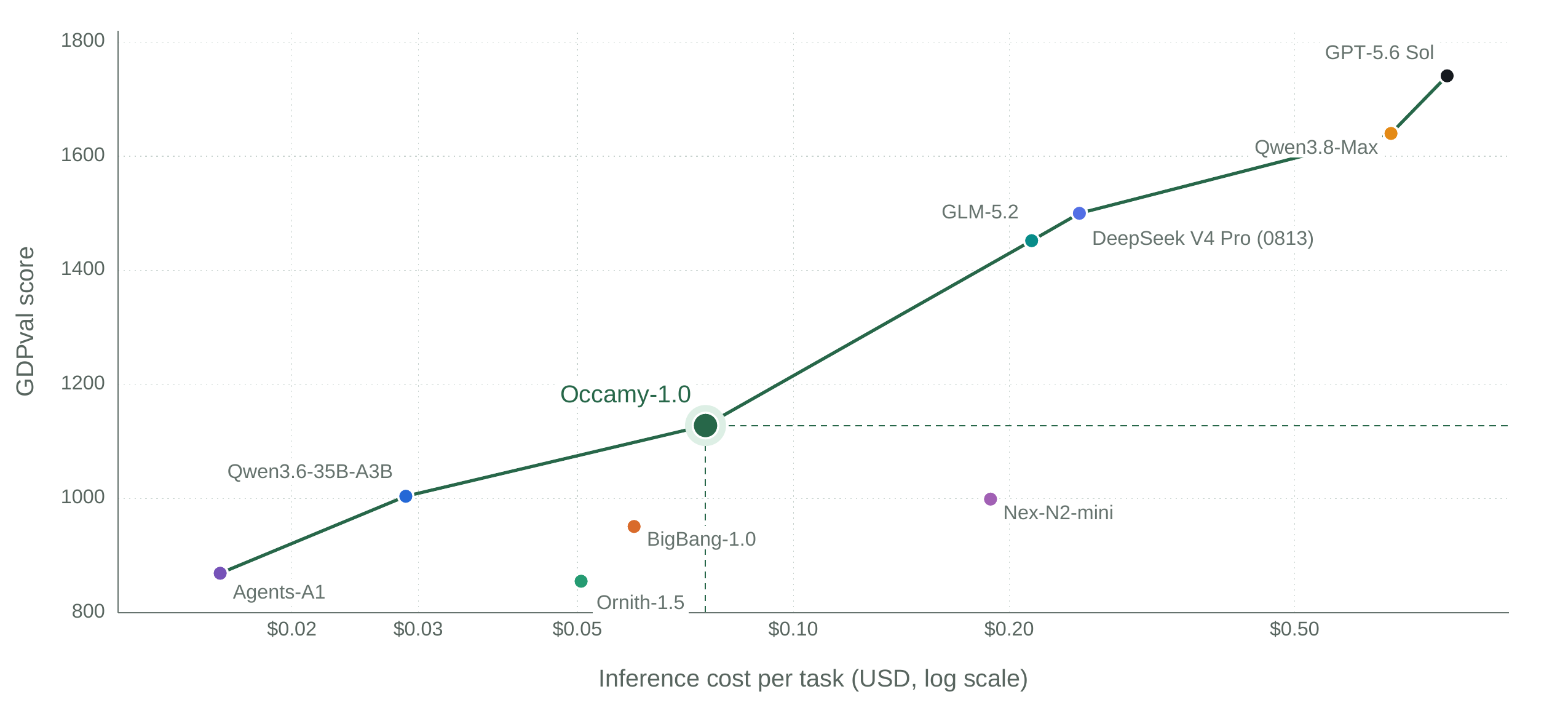}
\caption{GDPval cost--performance frontier (207-task frozen cohort).}
\label{fig:cost-gdpval}
\end{figure}

Figures~\ref{fig:cost-claweval}--\ref{fig:cost-gdpval} separate the aggregate comparison in Figure~\ref{fig:aggregate-cost-performance} into its four constituent benchmarks. Each panel uses the benchmark's native score and divides its total inference cost only by the number of tasks in that benchmark; the token-accounting and provider-selection protocol is otherwise unchanged. Reported costs are measurements under this protocol rather than invariant properties of a model. Hosted API runs may vary with provider routing and exposed cache accounting, while local runs may vary with cache residency, scheduling, and execution dynamics. We therefore interpret small cost differences with appropriate tolerance and emphasize the overall cost--performance trend.

\paragraph{Pricing snapshot and proxy mapping.}
Table~\ref{tab:cost-pricing-snapshot} freezes the price vectors used in the
figures as of September 2, 2026. For a controlled comparison of models with
the same serving footprint, all locally evaluated 35B-A3B checkpoints use the
lowest complete 35B-A3B price vector available in the snapshot: the
Qwen3.6-35B-A3B Darkbloom route. This common vector is applied to Occamy-1.0,
Qwen3.6-35B-A3B, Agents-A1, Nex-N2-mini, BigBang-1.0, and Ornith-1.5. Fixing the
unit-price vector within this group removes provider-price variation from the
comparison and makes the resulting cost differences attributable to measured
token usage. It is a size-matched monetary proxy, not a claim about checkpoint lineage or direct API
availability. The remaining systems use their corresponding
model-specific price vectors. Temporary free tiers and promotional discounts
are excluded. For no-cache routes, the cache-read column is set equal to the
ordinary input rate.

\begin{table}[ht]
\centering
\small
\setlength{\tabcolsep}{4pt}
\caption{Pricing snapshot used for the cost comparison. Rates are USD per
million tokens.}
\label{tab:cost-pricing-snapshot}
\resizebox{\linewidth}{!}{%
\begin{tabular}{lllrrr}
\toprule
Evaluated model(s) & Pricing proxy & Selected provider & Input & Cache read & Output \\
\midrule
\shortstack[l]{Occamy-1.0; Qwen3.6-35B-A3B; Agents-A1;\\ Nex-N2-mini; BigBang-1.0; Ornith-1.5}
  & Qwen3.6-35B-A3B & Darkbloom (no-cache route) & 0.05 & 0.05 & 0.70 \\
GPT-5.6 Sol & GPT-5.6 Sol & OpenAI & 4.00 & 0.40 & 20.00 \\
Qwen3.8-Max & Qwen3.8-Max & Alibaba Cloud International & 2.00 & 0.25 & 6.00 \\
DeepSeek V4 Pro (0813) & DeepSeek V4 Pro (0813) & DeepSeek & 0.66 & 0.022 & 1.98 \\
GLM-5.2 & GLM-5.2 & DigitalOcean & 0.70 & 0.105 & 2.20 \\
\bottomrule
\end{tabular}%
}
\end{table}

Table~\ref{tab:cost-per-task-snapshot} reports the exact per-task values used
to place the points in the four benchmark panels and in the equal-weight
aggregate. The final column is the macro-average of the four benchmark-level
costs, not a task-count-weighted micro-average.

\begin{table}[ht]
\centering
\small
\setlength{\tabcolsep}{4pt}
\caption{Inference cost per task (USD) used in the September 2, 2026 figures.}
\label{tab:cost-per-task-snapshot}
\begin{tabular}{lrrrrr}
\toprule
Model & Claw-Eval & WildClawBench & AutomationBench & GDPval & Macro avg. \\
\midrule
Occamy-1.0          & 0.026332 & 0.096500 & 0.034608 & 0.075455 & 0.058224 \\
Qwen3.6-35B-A3B     & 0.037839 & 0.128500 & 0.028701 & 0.028826 & 0.055967 \\
Agents-A1           & 0.040413 & 0.083369 & 0.030943 & 0.015886 & 0.042653 \\
Nex-N2-mini         & 0.040583 & 0.080736 & 0.068760 & 0.188366 & 0.094612 \\
BigBang-1.0         & 0.052362 & 0.105000 & 0.060511 & 0.059988 & 0.069465 \\
Ornith-1.5          & 0.028710 & 0.045032 & 0.064752 & 0.050608 & 0.047275 \\
GPT-5.6 Sol         & 0.756332 & 0.833667 & 0.739125 & 0.815773 & 0.786224 \\
Qwen3.8-Max         & 0.319548 & 0.431667 & 0.218797 & 0.681253 & 0.412816 \\
DeepSeek V4 Pro (0813) & 0.120000 & 0.066000 & 0.058383 & 0.250542 & 0.123731 \\
GLM-5.2             & 0.159849 & 0.211667 & 0.072083 & 0.214946 & 0.164636 \\
\bottomrule
\end{tabular}
\end{table}

\subsection{Benchmark-specific Protocols}
\label{app:eval_registry}

\paragraph{Primary co-work and office evaluation.}
\begin{itemize}
    \item \textbf{Claw-Eval}~\citep{claweval2026}: We evaluate Claw-Eval on a full set spanning the tool-use (T*) and user-agent (C*) splits in both English and Chinese, using the AccioBench orchestrator driving a containerized claw-eval runner (v7 image with the framework, mock services, and per-task graders baked in). Each task executes in an isolated Aone-Sandbox cloud container (8 GB memory and the benchmark's official recommended execution timeout) with sandbox tools enabled, at 50 concurrent trials with one attempt per task; per-task rewards are written by the bundled graders, and full agent traces are retained for analysis. The agent-under-test accesses the model through an OpenAI-compatible multicloud gateway pinned via backend-routing headers to a local SGLang deployment (four tensor-parallel replicas behind a load-balancing router) under long-context, deterministic inference settings, with a served-model identity gate ensuring every trial runs against the intended checkpoint. Because agent-side connection starvation can silently depress scores, all runs additionally pass integrity checks for trial completeness and endpoint contamination before scoring. We report the average score and pass$^3$ as the final metrics.
    \item \textbf{WildClawBench}~\citep{wildclawbench2026}: We evaluate WildClawBench, including 60 hand-written, open-ended agent tasks across 6 categories, at a pinned upstream commit, using the AccioBench orchestrator with a thin per-task wrapper around the native harness. Each trial runs the upstream OpenClaw-style agent loop inside an isolated Aone-Sandbox container whose runner image bakes in the benchmark repository and all prepared workspace assets (video clips, SAM3 weights, extracted git fixtures), so that trials are hermetic and reproducible; web-dependent tasks access the internet through a managed browsing sidecar with bounded retry semantics. Local checkpoints are served with SGLang under deterministic long-context settings and exposed to the sandboxed agent via an OpenAI-compatible gateway, while API baselines are accessed through the same provider-side gateway, keeping the agent scaffold identical across models. Scoring uses each task's embedded automated checks — per-task Python graders, of which roughly two-thirds invoke an LLM judge (GPT-5.4) — with judge traffic transparently proxied to an internal Azure endpoint so that grading is identical and network-independent across runs; trials execute as three independent runs per task, with one attempt in each run and 12-way concurrency, and full agent trajectories are retained for analysis. We report avg@3 over the three runs as the final score.
    \item \textbf{CommerceAgentBench (CAB)}~\citep{realreplica2026}. We evaluate CAB, which contains long-horizon agents in high-fidelity, stateful, and reproducible replicas of real-world commerce and business services. Its evaluation comprises 107 tasks spanning browser interaction, command-line tools, file and document manipulation, and API/MCP workflows, with text-only, browser-capable, and vision-required settings. Each task runs in a fresh isolated container, requires the agent to modify environment state or produce verifiable artifacts, and is scored using task-specific deterministic or LLM-assisted verifiers, thereby measuring end-to-end workflow completion rather than answer generation alone. We select OpenClaw as the harness for problem solving, and all other settings follow the benchmark's default configuration. We report pass$^1$ as the final score.
\item \textbf{GDPval}~\citep{gdpval2025}: We evaluate on the public
220-task GDPval release from OpenAI on Hugging Face. We independently select a
frozen cohort of 207 tasks, excluding 13 tasks with transport or
evaluation-integrity limitations in our runtime. This cohort is distinct from
GDPval-AA v2: Artificial Analysis additionally curated and revised the
benchmark tasks, but those task-level modifications are not publicly
available. We reproduce the accessible parts of its evaluation setup, running
each task once with Stirrup 0.1.12 in an isolated sandbox, the published
GDPval-AA v2 agent system prompt, the original public task prompt and reference
files, and the same context-window configuration, tool interface, 250-turn
limit, and 70\% context-compaction threshold. Submitted artifacts are evaluated
through blind pairwise comparisons against available frozen model and
human-reference submissions. Since Claude judge access was unavailable in our
environment, we use a deterministically balanced two-judge panel comprising
GPT-5.6-sol with \texttt{xhigh} reasoning effort and Gemini 3.1 Pro Preview
with \texttt{HIGH} thinking. We independently implement the disclosed
Bradley--Terry maximum-likelihood aggregation, treating ties as half-wins and
anchoring human performance at Elo 1000. We report the resulting score as
GDPval, an independent evaluation on the public GDPval release, not a
replication of GDPval-AA v2 or an official Artificial Analysis leaderboard
result.
    \item \textbf{Business Arena}~\citep{businessarena2026} evaluates end-to-end business operation in a dynamic cross-border marketplace. Starting with \$80{,}000, an agent must research the market, source and price products, manage inventory and capital, serve customers, respond to competition, and maintain compliance over a long horizon. Final net worth measures how much money the agent gains or loses through its operation. We follow the benchmark's original evaluation protocol, run five independent trials using the same seed, and report their average final net worth.
    \item \textbf{OfficeQA Pro}~\citep{officeqapro2026} evaluates long-document question
    answering over a corpus of U.S.\ Treasury Bulletin filings, where the agent is given
    a bash tool and must locate the answer by searching the corpus itself rather than
    being handed retrieved passages. Questions target specific figures that are
    scattered across filings and frequently must be reconciled across reporting periods,
    so answering requires the agent to plan a sequence of searches, narrow down the
    relevant tables, and carry out arithmetic over what it retrieves; individual filings
    reach 1.3\,MB (roughly 325k tokens), so reading a document end to end is not a
    viable strategy and the benchmark measures search competence rather than long-context
    recall. Scoring is exact match against the reference figure and is rule-based rather
    than judged, so the metric is deterministic given the model's outputs and a correct
    search that ends in the wrong value receives no partial credit. We evaluate the
    \texttt{officeqa\_pro} split (133 tasks) with one attempt per task and report
    accuracy over the split.
    \item \textbf{$\tau^3$-Bench (Banking)}~\citep{tau3bench2026}: We evaluate the
    \texttt{banking\_knowledge} domain of $\tau^3$-bench in full (all 97 tasks, no
    subsampling), in which an agent resolves customer-service requests against a
    ${\sim}700$-document knowledge base while conversing with a simulated user. We use
    the benchmark's \texttt{alltools} retrieval configuration, which exposes the agent
    to lexical BM25 search, dense embedding search, and a shell running in an isolated
    filesystem sandbox, so that retrieval is agent-driven rather than supplied by the
    harness. The agent under test runs at the highest available reasoning-effort tier,
    and GPT-5.2 with low reasoning effort acts as the user simulator. The retrieval
    configuration and the user simulator follow the official setting and are held fixed
    across all reported models. We report pass$^1$ as the final score.
\end{itemize}

\paragraph{Repository, terminal, and coding evaluation.}
\begin{itemize}
    \item \textbf{Terminal-Bench 2.1 (TB 2.1)}~\citep{terminalbench2026}: We evaluate TB 2.1, which is a benchmark of 89
verified tasks measuring an agent's ability to complete complex, real-world work
in a terminal, spanning software engineering (building, debugging, security
patching), data science, and scientific computing (e.g., protein assembly). Each task provides an isolated container environment defined by a
task-specific Docker image, a natural-language instruction, and a hidden
verification suite; the agent interacts with the environment exclusively by
issuing shell commands until it deems the task complete, after which the verifier
runs the task's test scripts inside the same environment. We evaluate with the
official Terminus~2 harness under the Harbor evaluation framework, serving the
model through an OpenAI-compatible endpoint; a task counts as resolved only if
its full verification suite passes, and we report the mean resolution rate over
all 89 tasks with one rollout per task.

\end{itemize}

\paragraph{Tool use and automation evaluation.}
\begin{itemize}
    \item \textbf{AutomationBench 1.0.6}~\citep{automationbench2026}: We evaluate on AutomationBench 1.0.6, for AI agents on cross-application workflow orchestration. The checkout is hash-verified before every run and fails closed on any drift, results can't be silently overwritten, and aborted or empty runs are retried rather than scored as zeros. A full run is the six public domains, including 600 tasks, against either a local endpoint or a gateway-hosted model, so different arms are directly comparable. Scoring is programmatic: each task has hidden assertions checked against the final simulated world state, with credit for freebies removed so a model can't score by doing nothing. Every run emits a receipt attesting to task count, pass/fail split, and completeness, and we only quote runs whose receipt shows a clean full 600 matching the other arms. We always report two numbers, including strict pass rate (all assertions must pass) as the headline and average partial-credit reward as the denser signal, because the strict metric alone hides most of the movement between checkpoints. 
    
    \item \textbf{BFCL v4}~\citep{bfcl2026}: We evaluate BFCL v4 (the full native benchmark: 22 subsets, 5,106 graded rows across 7 categories, with the official 10/10/10/30/40 weighting over non-live, live, irrelevance, multi-turn, and agentic) using evalscope 1.9.1 \cite{evalscope2024} driving a patched bfcl-eval 2025.10.27.1 harness. Local models are served with SGLang under long-context, deterministic inference settings, while API baselines are accessed through DashScope, with prediction trajectories retained for analysis. We introduce two important harness fixes: preserving final textual responses when no tool call is produced and replacing the quota-limited web-search backend with a more reliable service and bounded tool outputs. All runs additionally pass integrity checks for completeness and endpoint contamination before scoring. 
    
    \item \textbf{VitaBench}~\citep{vitabench2025}: We evaluate VitaBench on its complete
    400-task life-services benchmark, covering delivery, in-store, OTA, and
    cross-domain tracks under deterministic, standardized agent--simulator--tool
    interactions. DeepSeek V4 Flash serves as both the user simulator and the
    trajectory judge, with dedicated quota handling and sufficient reasoning budget,
    and is held fixed across all reported models; results are therefore comparable
    within this report but not directly to previously published VitaBench numbers.
    Local checkpoints are served through parallel SGLang replicas, while API models
    use their own provider configurations, and a run is valid only if it completes
    every task in all four tracks. We report the average judged reward across the
    four tracks as the final score.

\end{itemize}

\paragraph{Instruction following.}
\begin{itemize}

\item \textbf{IFEval}~\citep{zhou2023instruction}: We evaluate instruction-following ability on all 541 English prompts in the official IFEval split under a zero-shot setting. Each prompt contains one to three constraints drawn from 25 types of programmatically verifiable instructions, including length, keyword, formatting, punctuation, and response-structure requirements. We generate one response per prompt and evaluate it using the official deterministic checker. Generation uses temp=1.0, top\_p=0.95, top\_k=20, presence\_penalty=1.5, max\_tokens=32,768, and batch size 8.  We report the averaged strict accuracy at both the prompt and instruction levels: prompt-level accuracy requires all constraints in a prompt to be satisfied, whereas instruction-level accuracy scores each constraint separately. 

\end{itemize}

\section{Full Training Hyperparameters}
\label{app:training_hyperparameters}


\subsection{Supervised Fine-Tuning}
\label{sec:sft}

We perform full-parameter SFT of Qwen3.6-35B-A3B with the Megatron backend of
ms-swift 4.5; the full configuration is given in Table~\ref{tab:sft-config}.
The parallel layout is dictated by sequence length rather than model size: all
eight ranks are devoted to context parallelism, with expert parallelism folded
onto the same ranks. We set the packing length equal to the maximum sequence
length so that no trajectory is ever split across packs, since a trajectory
crossing a pack boundary would break the state dependencies the model is meant
to learn. Reasoning traces are retained in the targets, while empty
\texttt{think} spans are excluded from the loss to avoid learning degenerate
empty reasoning blocks. At this sequence length activation memory is the binding
constraint, so we recompute activations in full; we also disable NVLink-SHARP
multicast to avoid a known deadlock in the MoE all-to-all collective, which costs
little with a single data-parallel group.

\begin{table}[t]
\centering
\small
\caption{Detailed SFT configuration for Occamy. Extends Table~\ref{tab:sft_hparams}}
\label{tab:sft-config}
\begin{tabular}{ll}
\toprule
\textbf{Setting} & \textbf{Value} \\
\midrule
\multicolumn{2}{l}{\emph{Model and system}} \\
Initialization        & Qwen3.6-35B-A3B (full-parameter) \\
Framework             & ms-swift 4.5, Megatron backend \\
Hardware              & $1\times$ node, $8\times$ B200 \\
Parallelism           & TP\,1 / PP\,1 / CP\,8 / EP\,8 / ETP\,1 / DP\,1 \\
Precision             & bf16 \\
Attention             & FlashAttention \\
MoE kernels           & Grouped GEMM, permute fusion, shared-expert overlap \\
NVLink-SHARP          & Disabled (MoE all-to-all deadlock) \\
\midrule
\multicolumn{2}{l}{\emph{Data and sequence}} \\
Packing               & Bin-packed, padding-free \\
Sequence length       & $131{,}072$ \\
Loss masking          & \texttt{default+ignore\_empty\_think}, \texttt{preserve\_thinking} \\
\midrule
\multicolumn{2}{l}{\emph{Optimization}} \\
Optimizer             & Fused AdamW, bf16 moment states \\
$(\beta_1,\beta_2)$   & $(0.9,\,0.95)$ \\
Weight decay          & $0.1$ \\
Micro / global batch  & $1$ / $8$ packed sequences (grad.\ accum.\ $8$) \\
Peak learning rate    & $1\times10^{-5}$ \\
Schedule              & $5\%$ linear warmup, cosine decay to $2\times10^{-7}$ \\
MoE aux.\ loss        & $1\times10^{-3}$ \\
Epochs                & 5 \\
\bottomrule
\end{tabular}
\end{table}

\newcommand{\sao}{SAO}

\subsection{Reinforcement Learning with Context Compaction (\sao)}
\label{app:rl-pipeline}

Our final RL stage trains the merged checkpoint of two experts with a critic-based, segment-wise PPO over long-horizon agentic rollouts with \emph{trainable
context compaction}, following CompactionRL~\citep{li2026compactionrlreinforcementlearningcontext} and SAO \citep{hou2026singlerolloutasynchronousoptimizationagentic}. The full configuration is given in Table~\ref{tab:rl-hparams}. Rollouts execute real tool-use
tasks in sandboxed OpenClaw or Hermes sessions behind a recording proxy. When the
remaining context budget falls below the agent-framework threshold, the
\emph{policy itself} generates a summary of its history and the rollout resumes
from $\langle\text{system prompt},\ \text{resume template}(\text{summary}),\
\text{last }k\text{ steps}\rangle$; a rollout thus becomes $K$ segments
(execution and summarization), all sampled from the same trainable policy, all
optimized, and all sharing the single terminal task reward $R$: there is no
separate summary-quality reward. We use PPO with a learned critic initialized from the policy checkpoint, with critic-only value-pretraining steps and 2 critic
updates per policy update. Advantages use \emph{masked} per-segment GAE
computed over optimized tokens only (bootstrapping $V{=}0$ at the segment end,
never reading the critic at observation/tool positions, where the value loss is
masked), followed by the cross-trajectory correction
$\hat{A}_{s,i} = (\gamma\lambda_\tau)^{N_{>s}}\, A^{\mathrm{loc}}_{s,i}$,
where $N_{>s}$ counts optimized tokens in later segments of the same rollout,
so the composite discount $(\gamma\lambda_\tau)^{N_{>s}+n_s-i}$ matches each
token's true distance to the terminal reward in the concatenated rollout.
The GAE parameter is length-adaptive per trajectory \citep{yue2025vapoefficientreliablereinforcement},
$\lambda_\tau = 1 - \tfrac{1}{\alpha L_\tau}$ with $\alpha{=}1.5$ and $L_\tau$
as the trajectory's optimized-token count. The loss is normalized at the token
level over all optimized tokens in the batch, removing segment-count and
length bias. The KL coefficient is exactly zero. Training is fully asynchronous --- 2~rollout nodes
(16~GPUs) generate trajectories while 4~actor nodes (32~GPUs; TP$=2$, CP$=4$,
DP$=4$, EP$=8$, with the critic time-sharing the actor GPUs via optimizer
offload) train --- with double-sided importance sampling (DIS) for training stability. Our policy version staleness is 3, the agent context budget is set to
128k and the post-compaction context is ${\approx}\,0.45$ of the window.

\begin{table}[t]
\centering
\caption{Detailed SAO RL configuration for Occamy.}
\label{tab:rl-hparams}
\small
\begin{tabular}{ll}
\toprule
\textbf{Setting} & \textbf{Value} \\
\midrule
\multicolumn{2}{l}{\emph{Optimization}} \\
Policy learning rate            & $2\times10^{-6}$ \\
Critic learning rate            & $3\times10^{-6}$ \\
Optimizer                       & Adam ($\beta_1{=}0.9$, $\beta_2{=}0.98$) \\
Weight decay                    & $0.1$ \\
Gradient clipping               & $1.0$ \\
GAE $\gamma$                    & $1.0$ \\
GAE actor $\lambda_\tau$              & $1-\frac{1}{1.5\,L_\tau}$ \\
GAE critic $\lambda_\tau$              & $1.0$ \\
KL coefficient                  & $0$ \\
Entropy bonus                   & $0$ \\
DIS range                       & $0.8$ / $3.0$ \\
Loss normalization              & Token-level \\
\midrule
\multicolumn{2}{l}{\emph{Rollout}} \\
Trajectories per step  & $128$ (group size $1$) \\
Sampling temperature   & $0.7$ \\
Context budget         & $128{,}000$ \\
Per-decode cap         & $16{,}384$ tokens \\
Max segment length      & $90{,}000$ tokens \\
Compaction ratio       & $0.45$ \\
Actor training steps               & $100$ \\
\midrule
\multicolumn{2}{l}{\emph{Critic}} \\
Value-pretraining steps          & $20$ \\
Critic updates per policy update & $2$ \\
Initialization                   & Policy checkpoint \\
Micro-batching                   & Dynamic, $32{,}768$ tokens/GPU \\
\bottomrule
\end{tabular}
\end{table}

\newpage

\subsection{Model-Merge Implementation and Validation}
\label{app:merge_implementation}
The two parent checkpoints use the same architecture, tokenizer, and lineage, but may serialize equivalent MoE tensors under different layouts. Before averaging, we normalize per-expert and fused representations into a common layout and validate the gate, up, and down projection correspondences using cosine similarity. A merge is aborted when the expected alignment is weak or ambiguous. We accumulate floating-point averages in \texttt{float32}, cast them back to the parent dtype, inherit one-sided tensors such as the MTP draft head without averaging, and preserve the selected parent checkpoint's index, configuration, and tokenizer files. After writing, we independently recompute sampled dense tensors and expert slices and require exact agreement with the stored checkpoint. \section{Case Study}
\label{app:accio-work-case-study}
\definecolor{AWGreen}{HTML}{276749}
\definecolor{AWGreenBg}{HTML}{F1F7F3}
\definecolor{AWBlue}{HTML}{315B7D}
\definecolor{AWBlueBg}{HTML}{F2F6FA}
\definecolor{AWAmber}{HTML}{9A6700}
\definecolor{AWAmberBg}{HTML}{FFF8E8}
\definecolor{AWRed}{HTML}{A13D3D}
\definecolor{AWRedBg}{HTML}{FBF3F2}
\definecolor{AWInk}{HTML}{20242A}
\definecolor{AWMuted}{HTML}{667085}
\definecolor{AWRule}{HTML}{D9DEE7}

\subsection{Protocol and Evidence}
\label{sec:accio-work-cases}

\paragraph{Setup.}
We compare the final Occamy-1.0 checkpoint with the Qwen3.6-35B-A3B
starting checkpoint on three Accio Work workflows: project-state reconciliation,
PDF digestion, and dependency-safe incident recovery. The harness, fixtures,
tools, output contract, context, and decoding settings are held fixed; only the
served-model endpoint changes, and all three Occamy runs use the same final
checkpoint.

\begin{figure}[H]
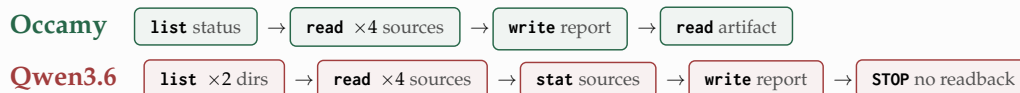

\centering
\begingroup
\newcommand{\tracecall}[3]{%
  \tcbox[on line,colback=#1!7,colframe=#1,boxrule=0.55pt,arc=0.7mm,
    left=1.0mm,right=1.0mm,top=0.48mm,bottom=0.48mm,nobeforeafter]
    {\scriptsize\textbf{\texttt{#2}}\;{\color{black!68}#3}}%
}
\newcommand{\tracearrow}{\hspace{0.35mm}{\scriptsize$\rightarrow$}\hspace{0.35mm}}

\begin{tcolorbox}[width=\textwidth,colback=white,colframe=AWRule,
  boxrule=0.55pt,arc=1.1mm,left=1.8mm,right=1.8mm,top=1.1mm,bottom=1.1mm]
\small\textbf{Project-state reconciliation.}
Read four records for projects Alpha, Beta, and Gamma; compute exact status
counts, preserve dependency lineage, distinguish ownership roles, and persist
an auditable report.
\end{tcolorbox}
\vspace{0.7mm}
\begin{minipage}[t]{0.492\textwidth}
\begin{tcolorbox}[equal height group=awproject,colback=AWGreenBg,colframe=AWGreen,
  boxrule=0.75pt,arc=1.1mm,left=1.7mm,right=1.7mm,top=1.1mm,bottom=1.1mm]
\textcolor{AWGreen}{\textbf{Occamy}}\hfill
\textcolor{AWGreen}{\textbf{VERIFIED}}

\small Correctly reported Beta as \textbf{1 completed / 2 active / 1 blocked /
4 overdue}. It preserved both source-level chains:
\texttt{609}$\rightarrow$\texttt{607} and
\texttt{606}$\rightarrow$\texttt{608}, then persisted and read back the report.

\end{tcolorbox}
\end{minipage}\hfill
\begin{minipage}[t]{0.492\textwidth}
\begin{tcolorbox}[equal height group=awproject,colback=AWRedBg,colframe=AWRed,
  boxrule=0.75pt,arc=1.1mm,left=1.7mm,right=1.7mm,top=1.1mm,bottom=1.1mm]
\textcolor{AWRed}{\textbf{Qwen3.6-35B-A3B}}\hfill
\textcolor{AWRed}{\textbf{COUNT ERROR}}

\small Reported only \textbf{3} overdue Beta tasks although 606--609 were all
unfinished and past due. It also attached an unsupported second blocker to
608; the source attaches the vendor-document blockage to 607. It persisted the
report but stopped without reading the artifact back.

\end{tcolorbox}
\end{minipage}

\vspace{1.0mm}
\begin{tcolorbox}[width=\textwidth,colback=black!1,colframe=AWRule,
  boxrule=0.6pt,arc=1.0mm,left=1.8mm,right=1.8mm,top=1.2mm,bottom=1.2mm,
  title={\small\textbf{Observed tool trajectory}},
  coltitle=AWInk,colbacktitle=black!4]
\textcolor{AWGreen}{\textbf{Occamy}}\quad
\tracecall{AWGreen}{list}{status}
\tracearrow\tracecall{AWGreen}{read $\times 4$}{sources}
\tracearrow\tracecall{AWGreen}{write}{report}
\tracearrow\tracecall{AWGreen}{read}{artifact}

\vspace{0.8mm}\par
\textcolor{AWRed}{\textbf{Qwen3.6}}\quad
\tracecall{AWRed}{list $\times 2$}{dirs}
\tracearrow\tracecall{AWRed}{read $\times 4$}{sources}
\tracearrow\tracecall{AWRed}{stat}{sources}
\tracearrow\tracecall{AWRed}{write}{report}
\tracearrow\tracecall{AWRed}{STOP}{no readback}
\end{tcolorbox}
\endgroup
\caption{Project-state reconciliation through the harness. The
upper cards compare verified artifacts; the lower panel shows the observed
tool trajectory for each endpoint. Occamy preserves exact counts and verifies
the persisted report, while Qwen3.6 undercounts and stops without readback.}
\label{fig:accio-work-project-state}
\end{figure}

Figures~\ref{fig:accio-work-project-state},
\ref{fig:accio-work-pdf-digest}, and
\ref{fig:accio-work-incident-recovery} pair checks of the persisted artifacts
with the chronological tool calls that produced them. Each task is run once per
endpoint, so these cases complement the aggregate evaluation rather than estimate
pass probability.

\par

\begin{figure}[H]
\centering
\begingroup
\newcommand{\tracecall}[3]{%
  \tcbox[on line,colback=#1!7,colframe=#1,boxrule=0.55pt,arc=0.7mm,
    left=1.0mm,right=1.0mm,top=0.48mm,bottom=0.48mm,nobeforeafter]
    {\scriptsize\textbf{\texttt{#2}}\;{\color{black!68}#3}}%
}
\newcommand{\tracearrow}{\hspace{0.35mm}{\scriptsize$\rightarrow$}\hspace{0.35mm}}

\begin{tcolorbox}[width=\textwidth,colback=white,colframe=AWRule,
  boxrule=0.55pt,arc=1.1mm,left=1.8mm,right=1.8mm,top=1.1mm,bottom=1.1mm]
\small\textbf{Twelve-document PDF digest.}
Recover titles and categories without mutating the PDFs, propose unique
collision-safe names, identify every caption-related paper, quote exact
Table~1 evidence, and reconcile all inputs and outputs.
\end{tcolorbox}
\vspace{0.7mm}
\begin{minipage}[t]{0.492\textwidth}
\begin{tcolorbox}[equal height group=awpdf,colback=AWGreenBg,colframe=AWGreen,
  boxrule=0.75pt,arc=1.1mm,left=1.7mm,right=1.7mm,top=1.1mm,bottom=1.1mm]
\textcolor{AWGreen}{\textbf{Occamy}}\hfill
\textcolor{AWGreen}{\textbf{FULL CONTRACT}}

\small Reconciled 12/12 PDFs into six categories and recovered the exact
caption set \{01,02,03,08\}. For each selected paper it preserved concrete
Table~1 evidence---dataset, method, and score---rather than relying on the
title alone. It detected both duplicate-title groups and generated readable,
deterministic names by combining a title slug with the source stem, e.g.,
\texttt{reliable\_image\_captioning\_paper\_01.pdf} and
\texttt{\ldots\_paper\_02.pdf}. All originals remained read-only; the manifest,
evidence table, and summary were written and read back.

\end{tcolorbox}
\end{minipage}\hfill
\begin{minipage}[t]{0.492\textwidth}
\begin{tcolorbox}[equal height group=awpdf,colback=AWAmberBg,colframe=AWAmber,
  boxrule=0.75pt,arc=1.1mm,left=1.7mm,right=1.7mm,top=1.1mm,bottom=1.1mm]
\textcolor{AWAmber}{\textbf{Qwen3.6-35B-A3B}}\hfill
\textcolor{AWAmber}{\textbf{WEAK NAMING}}

\small Also recovered 12/12 titles, the six-category census, and the exact
caption set, which isolates the defect from basic extraction. However, it
retained opaque input stems and appended \texttt{-1} to later collisions, e.g.,
\texttt{paper\_02-1.pdf}. The names are unique but omit the recovered-title
semantics requested by the task, and the trajectory ends after writing rather
than with an explicit output readback.

\end{tcolorbox}
\end{minipage}

\vspace{1.0mm}
\begin{tcolorbox}[width=\textwidth,colback=black!1,colframe=AWRule,
  boxrule=0.6pt,arc=1.0mm,left=1.8mm,right=1.8mm,top=1.2mm,bottom=1.2mm,
  title={\small\textbf{Observed tool trajectory}},
  coltitle=AWInk,colbacktitle=black!4]
\textcolor{AWGreen}{\textbf{Occamy}}\quad
\tracecall{AWGreen}{read $\times 6$}{parse error}
\tracearrow\tracecall{AWAmber}{discover}{pdftotext}
\tracearrow\tracecall{AWGreen}{extract}{12 PDFs}
\tracearrow\tracecall{AWGreen}{write $\times 3$}{artifacts}
\tracearrow\tracecall{AWGreen}{verify}{outputs}

\vspace{0.8mm}\par
\textcolor{AWRed}{\textbf{Qwen3.6}}\quad
\tracecall{AWRed}{read $\times 12$}{parse error}
\tracearrow\tracecall{AWAmber}{discover}{pdftotext}
\tracearrow\tracecall{AWAmber}{extract}{no match}
\tracearrow\tracecall{AWRed}{repair}{pattern}
\tracearrow\tracecall{AWRed}{write $\times 3$}{artifacts}
\end{tcolorbox}

\vspace{0.7mm}
{\scriptsize\textcolor{AWMuted}{Amber marks a failed path or recovery call.
Tool names and order come from the harness records.}}
\endgroup
\caption{Twelve-document PDF digestion through the harness. Both
agents recover from parser failure, but Occamy produces title-derived,
collision-safe names and verifies its outputs; Qwen3.6 requires another repair
and retains opaque input stems. The lower panel exposes the full comparison
trajectory rather than reporting only the final files.}
\label{fig:accio-work-pdf-digest}
\end{figure}

\par

\begin{figure}[H]
\centering
\begingroup
\newcommand{\tracecall}[3]{%
  \tcbox[on line,colback=#1!7,colframe=#1,boxrule=0.55pt,arc=0.7mm,
    left=1.0mm,right=1.0mm,top=0.48mm,bottom=0.48mm,nobeforeafter]
    {\scriptsize\textbf{\texttt{#2}}\;{\color{black!68}#3}}%
}
\newcommand{\tracearrow}{\hspace{0.35mm}{\scriptsize$\rightarrow$}\hspace{0.35mm}}

\begin{tcolorbox}[width=\textwidth,colback=white,colframe=AWRule,
  boxrule=0.55pt,arc=1.1mm,left=1.8mm,right=1.8mm,top=1.1mm,bottom=1.1mm]
\small\textbf{Dependency-safe incident recovery.}
Read five operational records, identify one root cause and its downstream
propagation, reconcile four inventory items, and persist a recovery plan whose
ordering prevents stale or partially migrated data from triggering restock.
\end{tcolorbox}
\vspace{0.7mm}
\begin{minipage}[t]{0.492\textwidth}
\begin{tcolorbox}[equal height group=awincident,colback=AWGreenBg,colframe=AWGreen,
  boxrule=0.75pt,arc=1.1mm,left=1.7mm,right=1.7mm,top=1.1mm,bottom=1.1mm]
\textcolor{AWGreen}{\textbf{Occamy}}\hfill
\textcolor{AWGreen}{\textbf{SAFE ORDER}}

\small Identified the retired v2 supplier API as the single root cause and
preserved the chain \texttt{JOB-1501}$\rightarrow$\texttt{1502}
$\rightarrow$\texttt{1503}. Its plan follows the required safety order:
\textbf{backup $\rightarrow$ activate v3 $\rightarrow$ configure
JSON/cursor $\rightarrow$ connectivity check $\rightarrow$ deactivate v2
$\rightarrow$ full sync $\rightarrow$ reconcile $\rightarrow$ restock
$\rightarrow$ notify}.

\end{tcolorbox}
\end{minipage}\hfill
\begin{minipage}[t]{0.492\textwidth}
\begin{tcolorbox}[equal height group=awincident,colback=AWRedBg,colframe=AWRed,
  boxrule=0.75pt,arc=1.1mm,left=1.7mm,right=1.7mm,top=1.1mm,bottom=1.1mm]
\textcolor{AWRed}{\textbf{Qwen3.6-35B-A3B}}\hfill
\textcolor{AWRed}{\textbf{ORDER MISMATCH}}

\small Found the same root cause and dependency chain, but collapsed
activation, switching, and connectivity into one step; it then scheduled the
full synchronization while the legacy v2 integration was still active and
deferred deactivation until after downstream jobs and notifications. The
content is fluent, but the recovery sequence violates the task's safety gate.

\end{tcolorbox}
\end{minipage}

\vspace{1.0mm}
\begin{tcolorbox}[width=\textwidth,colback=black!1,colframe=AWRule,
  boxrule=0.6pt,arc=1.0mm,left=1.8mm,right=1.8mm,top=1.2mm,bottom=1.2mm,
  title={\small\textbf{Observed tool trajectory}},
  coltitle=AWInk,colbacktitle=black!4]
\textcolor{AWGreen}{\textbf{Occamy}}\quad
\tracecall{AWGreen}{read $\times 5$}{records}
\tracearrow\tracecall{AWAmber}{write}{arg error}
\tracearrow\tracecall{AWGreen}{re-read $\times 5$}{records}
\tracearrow\tracecall{AWGreen}{write}{brief}
\tracearrow\tracecall{AWGreen}{read}{artifact}

\vspace{0.8mm}\par
\textcolor{AWRed}{\textbf{Qwen3.6}}\quad
\tracecall{AWRed}{read $\times 5$}{records}
\tracearrow\tracecall{AWAmber}{write}{arg error}
\tracearrow\tracecall{AWRed}{re-read $\times 5$}{records}
\tracearrow\tracecall{AWRed}{write}{brief}
\tracearrow\tracecall{AWRed}{read}{artifact}
\end{tcolorbox}

\vspace{0.7mm}
{\scriptsize\textcolor{AWMuted}{Both lanes used the same five input bytes,
tool surface, decoding settings, and recovery opportunity. The
comparison grades the persisted plan against a predeclared operational safety
order rather than against writing style.}}
\endgroup
\caption{Dependency-safe incident recovery through the harness.
Both agents identify the incident, recover from the same write-call failure,
and persist a report; only Occamy preserves the required migration and
downstream-execution order.}
\label{fig:accio-work-incident-recovery}
\end{figure}

\par

\subsection{Interpretation and Limitations}Across the three controlled workflows, Occamy's advantage appears in exactness, artifact verification, and dependency-safe sequencing rather than prose style alone. The evidence is qualitative: each task is run once per endpoint under fixed fixtures and the same Accio Work harness. These examples therefore illustrate execution behavior and failure modes; they do not replace the aggregate evaluation in Section~
6 or establish robustness to sampling, alternative task formulations, or broader tool surfaces.

\section{Detailed Co-work Episode Anatomy}
\label{app:episode-anatomy}
\label{sec:history-rewrites}Figure~\ref{fig:episode-anatomy} expands the boundaries summarized in Table~\ref{tab:cowork-terminology}. It shows how delegation branches one task-scoped episode into multiple runs and trajectories, while a history rewrite divides an individual trajectory into append-only segments without resetting the shared environment or task-level outcome.\begin{center}\begin{minipage}{\linewidth}\centering\includegraphics[width=\linewidth]{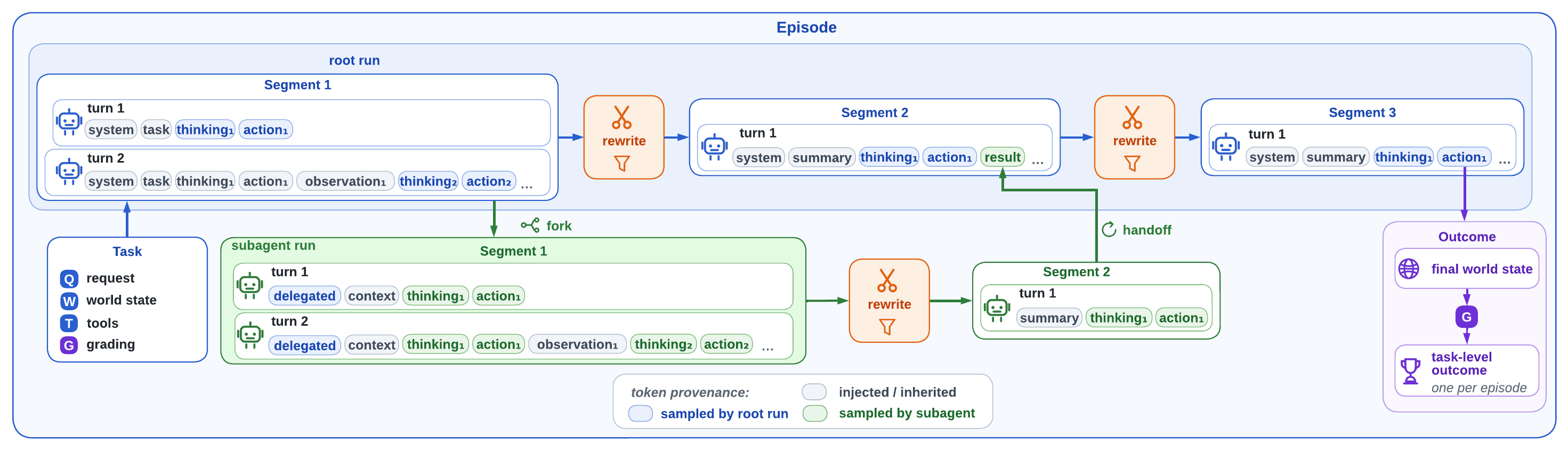}\captionsetup{hypcap=false}\captionof{figure}{Anatomy of a co-work episode. A task starts one episode. The root run coordinates the work and delegates a subgoal to a subagent run, whose result returns to the parent as an observation. Each run is recorded as its own trajectory. Within a run, a history rewrite closes the current append-only segment and starts the next from the rewritten prefix, while the run, shared environment, and episode continue. The grader produces one task-level outcome for the complete episode.}\label{fig:episode-anatomy}\end{minipage}\end{center}

\end{document}